\pdfoutput=1

\documentclass{article}
\usepackage{ijcai19}

\usepackage{times}
\usepackage{soul}
\usepackage{url}
\usepackage{hyperref}
\usepackage[utf8]{inputenc}
\usepackage[small]{caption}
\usepackage{graphicx}
\usepackage{subfigure}
\usepackage{amsmath, nccmath}
\usepackage{mathrsfs}
\usepackage{amssymb}

\usepackage{booktabs}
\usepackage{algorithm}
\usepackage{algorithmicx} 
\usepackage[noend]{algpseudocode}
\title{Gradient Boosting with Piece-Wise Linear Regression Trees}
\author{
Yu Shi
\and
Jian Li\And
Zhize Li
\affiliations
Institute for Interdisciplinary Information Sciences, Tsinghua University, Beijing, China\\
\emails
shiyu17@mails.tsinghua.edu.cn,
lijian83@mail.tsinghua.edu.cn,
zz-li14@mails.tsinghua.edu.cn
}
\begin{document}
\maketitle
\begin{abstract}
Gradient Boosted Decision Trees (GBDT) is a very successful ensemble learning algorithm widely used across a variety of applications. 
Recently, several variants of GBDT training algorithms and implementations have been designed and heavily optimized in some very popular open sourced toolkits including XGBoost, LightGBM and CatBoost. 
In this paper, we show that both the accuracy and efficiency of GBDT can be further enhanced by using more complex base learners. Specifically, we extend gradient boosting to use \textit{piecewise linear regression trees} (PL Trees), instead of \textit{piecewise constant regression trees}, as base learners. We show that PL Trees can accelerate convergence of GBDT and improve the  accuracy. We also propose 
some optimization tricks to substantially reduce the training time of PL Trees, with little sacrifice of accuracy. 
Moreover, we propose several implementation techniques to speedup our algorithm on modern computer architectures with powerful Single Instruction Multiple Data (SIMD) parallelism. 
The experimental results show that GBDT with PL Trees can provide
very competitive testing accuracy with comparable or less training time. 
\end{abstract}
\section{Introduction}
\noindent Gradient Boosted Decision Trees (GBDT) \cite{friedman2001greedy} has shown its excellent performance in many real world applications and data science competitions \cite{tyree2011parallel,chen2012combining}. Decision trees widely used as base learners by GBDT assign a single predicted value for data on the same leaf. 
We call these decision trees \textit{piecewise constant regression trees}, since each tree defines a piecewise constant function in the input space. ID3 \cite{quinlan1986induction}, C4.5 \cite{quinlan2014c4} and CART \cite{breiman2017classification} are famous algorithms for training standalone piecewise constant decision trees. \cite{tyree2011parallel,chen2016xgboost} propose efficient algorithms for training them as base learners of GBDT. It is very likely that with more complex decision tree model, we can enhance the power of gradient boosting algorithms. The most natural extension to piecewise constant trees is replacing the constant values at the leaves by linear functions, so called \textit{piecewise linear regression trees} (PL Trees). This idea has been explored in \cite{wang2014boosted,hall2009weka,kuhn2012cubist}. However, due to its heavy computation cost, so far there's no fast and scalable implementation of gradient boosting with PL Trees. 

In this paper, we provide a fast and scalable implementation of gradient boosting with PL Trees. Our algorithm has training cost comparable to carefully optimized GBDT toolkits including XGBoost \cite{chen2016xgboost}, LightGBM \cite{ke2017lightgbm} and CatBoost \cite{prokhorenkova2018catboost}, all of which use piecewise constant trees as base learners. We reduce the cost of training PL Trees from both algorithmic and system aspects. From algorithmic level, we adopt an incremental feature selection strategy during the growth of a tree to constrain the size of linear models. The histogram technique (see e.g., \cite{tyree2011parallel,chen2016xgboost}) used by piecewise constant trees is also adapted to PL Trees. We then propose \textit{half-additive fitting} to further reduce the cost of fitting linear models. 
From system level, SIMD parallelism is very suitable for speeding up the training of PL Trees. However, cache must be efficiently utilized to provide operands fast enough for SIMD instructions. We arrange data structures carefully to reduce cache misses. All these techniques together make our algorithm more efficient than existing GBDT algorithms.   

The main contributions of our work are summarized as follows:
\begin{itemize}
\item We extend GBDT with second-order approximation to ones that use PL Trees as base learners. Our experiments demonstrate that PL Trees can improve the convergence rate of GBDT. 
\item We design an efficient strategy to fit the linear models in tree nodes, with incremental feature selection and half-additive fitting. This strategy avoids the 
prohibitive computational cost 
for fitting large linear models repeatedly when training a PL Tree. 
\item 
We propose several implementation techniques to exploit the power of SIMD parallelism by reducing cache misses in PL Tree training. 
\item We evaluate our algorithm on 10 public datasets, and compare it with state-of-the-art toolkits including XGBoost, LightGBM and CatBoost. The experimental results show that our algorithm can improve accuracy with comparable training time on numerical dense data. 
\end{itemize}

\section{Review of Gradient Boosted Decision Trees}  
In this section, we provide a brief review of GBDT. 
Specifically, we review one of the most popular variant XGBoost \cite{chen2016xgboost},
which uses second-order approximation of loss function \cite{friedman2000additive}. Second-order approximation is important for fast convergence of GBDT \cite{sun2014convergence}. 
Given a dataset $\mathcal{D}=\{(\mathbf{x}_i, y_i)\}_1^n$
with $m$ features, 
and $\mathbf{x}_i\in \mathbb{R}^m$, GBDT trains a sequence of decision trees $\{t_k\}_1^T$. The final output is the summation of these trees 
$\hat{y_i}=\sum_{k=1}^T t_k(\mathbf{x}_i).$
The loss function is usually augmented 
by regularization terms $\Omega{(t_k)}$ to prevent overfitting. $\Omega(t_k)$ reflects the complexity of tree $t_k$. Let $l:\mathbb{R}^2 \to \mathbb{R}$ be the loss function for a single data point. The total loss
$
\mathcal{L} = \sum_{i=1}^n l(\hat{y_i}, y_i) + \sum_{k=1}^T \Omega(t_k).
$
Let $\hat{y_i}^{(k)}$ be the predicted value of $\mathbf{x}_i$ after iteration $k$. At iteration $k+1$, a new tree $t_{k+1}$ is trained to minimize the following loss.
\begin{align*}
\mathcal{L}^{(k+1)} &= \sum_{i=1}^n l(\hat{y_i}^{(k+1)}, y_i) + \sum_{k'=1}^{k+1} \Omega(t_{k'})\\	
&= \sum_{i=1}^n l(\hat{y_i}^{(k)}+ t_{k+1}(\mathbf{x}_i), y_i) + \sum_{k'=1}^{k+1} \Omega(t_{k'})
\end{align*}
We can approximate the loss above.
$$
\mathcal{L}^{(k+1)}\approx \mathcal{C} + \Omega(t_{k+1}) + \sum_{i=1}^n 
\left[\frac{1}{2} h_i t_{k+1}(\mathbf{x}_i)^2 + g_i t_{k+1}(\mathbf{x}_i)\right]
$$
Here $\mathcal{C}$ is a constant value independent of $t_{k+1}$, $g_i = \frac{ \partial l(\hat{y_i}, y_i)}{\partial \hat{y_i}}|_{\hat{y_i}=\hat{y_i}^{(k)}}$ and $h_i = \frac{\partial^2 l(\hat{y_i}, y_i)}{\partial \hat{y_i} ^2}|_{\hat{y_i}=\hat{y_i}^{(k)}}$. Leaving out the constant, we get the objective of iteration $k+1$. 
\begin{fleqn}
\begin{equation}
\widetilde{\mathcal{L}}^{(k+1)}=\Omega(t_{k+1}) + \sum_{i=1}^n 
\left[\frac{1}{2} h_i t_{k+1}(\mathbf{x}_i)^2 + g_i t_{k+1}(\mathbf{x}_i)\right] 
\label{loss}
\end{equation}
\end{fleqn}
The specific form of regularizer $\Omega$ varies with the type of base learner.  
\section {Gradient Boosting with PL Trees}   	
In this section, we derive GBDT with second-order approximation using PL Trees. 
Formally, there are two basic components of our PL Trees,	
\begin{itemize}	
\item \textbf{Splits}: A split associated with an internal node is a condition used to partition the data in the node to its two child nodes.
Our PL Trees use univariate splits in the form $\mathbf{x}_{i,j} \le c$, where $\mathbf{x}_{i,j}$ is the $j$th feature value of data point $\mathbf{x}_i \in \mathbb{R}^m$. The feature $j$ is called the 
\textit{split feature}.  
\item \textbf{Linear Models}: On each leaf $s$, there is a linear model $f_s(\mathbf{x}_{i})=b_s + \sum_{j=1}^{m_s} \alpha_{s,j} \mathbf{x}_{i, k_{s,j}}$, where $\{\mathbf{x}_{i, k_{s,j}}\}_{j=1}^{m_s}$ is a subset of $\{\mathbf{x}_{i,j}\}_{j=1}^{m}$. We call features $\{k_{s,j}\}_{j=1}^{m_s}$ the \textit{regressors} for leaf $s$. 
The selection of the regressors 
is described in Section 4. 
\end{itemize}

Starting from a single root node, a PL Tree is trained by greedily splitting
nodes into children until the number of leaves in the tree reaches a preset maximum value. To give a clear framework for the training of PL Trees, we first define two operations:	
\begin{enumerate}
\item $\textit{FitNode}(s)$ fits a linear function on data in leaf $s$. The parameters of the function are calculated analytically to minimize (\ref{loss}). 	
\item $\textit{SplitEval}(s,j,c)$. For a leaf $s$ in tree $t_{k+1}$ of (\ref{loss}), a variable $j$ and a real value $c$, it returns the reduction of $\widetilde{\mathcal{L}}^{(k+1)}$, when splitting leaf $s$ with $\mathbf{x}_{i,j} \le c$ and fitting data in both
child nodes using \textit{FitNode}.  
 \end{enumerate}
Now the framework for training a PL Tree is summarized in Algorithm 1. We will spell out the details for \textit{FitNode} and \textit{SplitEval} later in this section.
\begin{algorithm}[t]
    \caption{Training Process of PL Tree} 	
        \begin{algorithmic}[1]
                 \State{initialize the tree with a single root node} 
        		\State{put all the sample points in root node}
        		\While{number of leaves fewer than a preset value}  	
			\For{each leaf $s$}
				\State{$j^*_s,c^*_s \gets \textrm{argmax}_{j,c} \textit{SplitEval(s,j,c)}$}	
			\EndFor 
			\State{$\hat{s} \gets \textrm{argmax}_s {\textit{SplitEval}(s,j_s^*, c_s^*)} $}
				\State{split $\hat{s}$ with condition $\mathbf{x}_{i,j^*_{\hat{s}}}\le c^*_{\hat{s}}$ into $s_1$ and $s_2$ }
				\State{$\textit{FitNode}(s_1)$, $\textit{FitNode}(s_2)$} 
		\EndWhile 
        \end{algorithmic}
\end{algorithm}
Let $\mathcal{I}_s$ be the set of data in leaf $s$ of tree $t_{k+1}$ in (\ref{loss}). We can rewrite (\ref{loss}) as follows.
$$
\widetilde{\mathcal{L}}^{(k+1)} =\Omega(t_{k+1}) + \sum_{s}\sum_{i\in \mathcal{I}_s} 
\left[\frac{1}{2} h_i t_{k+1}(\mathbf{x}_i)^2 + g_i t_{k+1}(\mathbf{x}_i)\right]
$$
Let $f_s$ be the linear model fitted in leaf $s$.  	
We use regularization term 
$
\Omega(t_k) = \lambda \sum_{s\in t_{k+1}} \omega(f_s). 	
$
Here $\omega(f_s)$ is the $L^2$ norm of parameters of linear model in leaf $s$. This prevents the linear models in the leaves from being too steep.	
Leaving out the $k$ notation and focusing on the loss of a single leaf $s$.
\begin{equation}
\widetilde{\mathcal{L}}_s= \omega(f_s) + \sum_{i\in \mathcal{I}_s} 
\left[\frac{1}{2}h_i f_s(\mathbf{x}_i)^2 + g_i f_s(\mathbf{x}_i) \right]
\label{ls}
\end{equation}

We first focus on fitting an optimal linear model for leaf $s$ given regressors $\{{k_{s,j}}\}_{j=1}^{m_s}$. The choice of regressors $\{{k_{s,j}}\}_{j=1}^{m_s}$ is left to Section 4.  
Let $\alpha_s=[b_s, \alpha_{s,1},...,\alpha_{s,m_s}]^T$.	
Substituting  $f_s(\mathbf{x}_i)$ into (\ref{ls}), we get the loss in terms of $\alpha_s$.
\begin{equation*}	
\begin{split}
\widetilde{\mathcal{L}}_s=&\sum_{i\in \mathcal{I}_s} 
\Bigl[\frac{1}{2}h_i (b_s + \sum_{j=1} ^{m_s} \alpha_{s,j} \mathbf{x}_{i, k_{s,j}})^2 \\
& + g_i(b_s + \sum_{j=1} ^{m_s} \alpha_{s,j} \mathbf{x}_{i,  k_{s,j}}) \Bigr] 
+ \frac{\lambda}{2} \|\alpha_s\|_2^2 
\end{split}
\end{equation*}	
Let $\mathbf{H}=\textrm{diag}(h_1,..., h_n)$, $\mathbf{g}=[g_1, ...,g_n]^T$, and $\mathbf{H}_s$ and $\mathbf{g}_s$ be the submatrix and subvector of $\mathbf{H}$ and $\mathbf{g}$ respectively by selecting $h_i$ and $g_i$ for $i\in \mathcal{I}_s$. Let $\mathbf{X}_s$ be the matrix of data in $\mathcal{I}_s$ with features $\{ k_{s,j} \}_{j=1}^{m_s}$, augmented by a column of 1's. 
We can write the loss $\widetilde{\mathcal{L}}_s$ in a concise form:
\begin{equation*}
\widetilde{\mathcal{L}}_s = \frac{1}{2} {\alpha_s}^T (\mathbf{X}_s^T \mathbf{H}_s \mathbf{X}_s + \lambda \mathbf{I}) \alpha_s + \mathbf{g}_s^T \mathbf{X}_s  \alpha_s.
\end{equation*}  
Thus the optimal value of $\alpha$ can be calculated analytically.
\begin{equation}
\alpha_s^* = -(\mathbf{X}_s^T \mathbf{H}_s \mathbf{X}_s + \lambda \mathbf{I})^{-1}\mathbf{X}_s^T \mathbf{g}_s
\label{alpha}
\end{equation}
Calculation of Equation (\ref{alpha}) is exactly $\textit{FitNode}(s)$. 
Then we get the minimum loss of leaf $s$. 
\begin{equation}
\widetilde{\mathcal{L}}_s^* = -\frac{1}{2} \mathbf{g}_s^T \mathbf{X}_s (\mathbf{X}_s^T \mathbf{H}_s \mathbf{X}_s + \lambda \mathbf{I})^{-1} \mathbf{X}_s^T \mathbf{g}_s
\label{minloss}
\end{equation}
When splitting a leaf $s$ into child $s_1$ and $s_2$ with condition $\mathbf{x}_{i,j} \le c$, we split the matrix $\mathbf{X}_s$ into sub-matrices $\mathbf{X}_{s_1}$ and $\mathbf{X}_{s_2}$ accordingly. 
Similarly we define $\mathbf{H}_{s_1}$, $\mathbf{H}_{s_2}$, $\mathbf{g}_{s_1}$ and $\mathbf{g}_{s_2}$. With these notations and the definition in (\ref{alpha}), the results of $\textit{FitNode}(s_1)$ and $\textit{FitNode}(s_2)$ are $\alpha_{s_1}^*$ and $\alpha_{s_2}^*$. Similarly we define $\widetilde{\mathcal{L}}^*_{s_1}$ and $\widetilde{\mathcal{L}}^*_{s_2}$ as in (\ref{minloss}). Then the reduction of loss incurred by
splitting $s$ into $s_1$ and $s_2$ is as follows. 
\begin{equation}
\textit{SplitEval}(s, j, c) = \widetilde{\mathcal{L}}_{s_1}^* + \widetilde{\mathcal{L}}_{s_2}^* - \widetilde{\mathcal{L}}_s^* 
\label{gain}
\end{equation} 
\section {Algorithmic Optimization}    	
In Algorithm 1, \textit{SplitEval} and \textit{FitNode} are executed repeatedly. For each candidate split, we need to calculate Equation (\ref{minloss}) twice for both child nodes. We use Intel MKL \cite{wang2014intel} to speedup the calculation, but it is still very expensive when the number of regressors is large. In this section, we introduce algorithmic optimizations to reduce the cost. 
\subsection{Histograms for GBDT with PL Trees} 
Histogram is an important technique to speedup GBDT  \cite{tyree2011parallel} by reducing the number of candidate splits. However, the construction of histograms becomes the most expensive part of tree training. We extend the histogram technique for PL Tree in GBDT. With piecewise constant trees, each bin in a histogram only needs to record the sum of gradients and hessians of data in that bin \cite{chen2016xgboost}. For PL Trees, the statistics in the histogram is more complex \cite{vogel2007scalable}. Two components in (\ref{minloss}) require summation over leaf data. 
\begin{equation}
\mathbf{X}_s^T \mathbf{H}_s \mathbf{X}_s = \sum_{i\in s} h_i \mathbf{x}_i \mathbf{x}_i^T, \quad \mathbf{X}_s^T \mathbf{g}_s = \sum_{i\in s} g_i \mathbf{x}_i 
\label{hist}
\end{equation} 
For simplicity, here we use $\mathbf{x}_i$ for the column vector of selected regressors of data $i$. Thus each bin $B$ needs to record both $\sum_{i\in B} h_i \mathbf{x}_i \mathbf{x}_i^T$ and $\sum_{i\in B} g_i \mathbf{x}_i$. And histogram construction becomes much more expensive. In Section 5, we introduce methods to speedup histogram construction. 
\subsection{Incremental Feature Selection and Half-Additive Fitting}  
It is unaffordable to use all features as regressors when fitting the linear models. For each node, we need to select a small subset of the features as regressors. 

In fact, the regressor selection can be done automatically as the tree grows \cite{friedman1979tree,vens2006simple}. Considering splitting $s$ into $s_1$ and $s_2$ with condition $\mathbf{x}_{i,q} \le c$, it is very natural to add split feature $q$ into the regressor sets of $s_1$ and $s_2$. The intuition is that, if this split should result in a significant reduction in the loss function, then feature $q$ contains relatively important information for the fitting of linear models in $s_1$ and $s_2$. Thus, for each leaf $s$, we choose the split features of its ancestor nodes as regressors. Formally, suppose the linear model of leaf $s$ is $f_s(\mathbf{x}_{i})=b_s + \sum_{j=1}^{m_s} \alpha_{s,j} \mathbf{x}_{i,k_{s,j}}$. Then the linear model of $s_1$ is $f_{s_1}(\mathbf{x}_{i})=b_{s_1} + \sum_{j=1}^{m_{s}} \alpha_{s_1,j} \mathbf{x}_{i,k_{s,j}} + \alpha_{s_1, m_{s} + 1}\mathbf{x}_{i,q}$. Similarly we have the linear model for $s_2$. When the number of regressors reach a preset threshold $d$, we stop adding new regressors to subsequent nodes. We call this \textit{incremental feature selection}. To decide the parameters in $f_{s_1}$, we have 3 approaches. 
\begin{enumerate}
\item \textit{additive fitting}: Only $b_{s_1}$ and $\alpha_{s_1, m_s + 1}$ are calculated. Coefficients of other regressors directly follow those of node $s$. This is the approach taken by \cite{friedman1979tree}. 
\item \textit{fully-corrective fitting}: All parameters of node $s_1$ are recalculated optimally according to (\ref{alpha}). 
\item \textit{half-additive fitting}: We have the following model of $s_1$. $$f_{s_1}(\mathbf{x}_i)=b_{s_1} + \beta \left( \sum_{j=1}^{m_s} \alpha_{s,j} \mathbf{x}_{i, k_{s,j}} \right) + \alpha_{s_1, m_s + 1} \mathbf{x}_{i,q}$$ Node $s_1$ takes the value $\sum_{j=1}^{m_s} \alpha_{s, j} \mathbf{x}_{i,k_{s,j}}$ as a combined regressor, and learns 3 parameters $b_{s_1}$, $\alpha_{s_1,m_s+1}$ and a scaling parameter $\beta$.
\end{enumerate}
Fully-corrective fitting provides the optimal parameters, while additive fitting has the lowest cost. Half-additive fitting combines the two and make a good trade-off between accuracy and efficiency, which is shown in Section 6.2. When adding the new regressor $q$ to node $s_1$, $\beta$ rescales the coefficients of regressors shared by parent node $s$. When the regressors are orthogonal and zero-mean in $s$ and $s_1$, and with square loss, it is easy to check that the 3 approaches produce the same result. 
\section{System Optimization}    	
In this section, we show how to speedup our algorithm, specifically the histogram construction, on modern CPUs. 
With slight abuse of notation, we use $n$ to denote the number of data points in the leaf, and $N$ to denote the size of training set. Each data point has a unique ID, ranging from $1$ to $N$. For each leaf $s$, an array $index_s$ of length $n$ is maintained to record these ID's for all data points in the leaf. For each feature $j$, we maintain an array $bin_j$ of length $N$. For a data point with unique ID $id$, $bin_j[id]$ records which bin the data point falls in the histogram of feature $j$. 
With these notations, we summarize the histogram construction process for leaf $s$ and feature $j$
in Algorithm 2. As mentioned in Section 4.1, the multiple terms $[h_i \mathbf{x}_i \mathbf{x}_i^T, g_i \mathbf{x}_i]$ making the histogram construction expensive. Next, we introduce two important techniques to speedup the construction. 
  \begin{algorithm}[H]
    \caption{Histogram Construction for Feature $j$ on Leaf $s$}
    \begin{algorithmic}[1] 
		\State{\textbf{Input}: $\mathbf{x}_1,...,\mathbf{x}_n$, $g_1,...,g_n$, $h_1,...,h_n$, $bin_j$, $index_s$}  
		\State{\textbf{Output}: histogram $hist_{j,s}$}
		\For{i = 1 to n}
			\State{id = $index_s$[i]}
			\State{bin = $bin_j$[id]}	
			\State{$hist_{j,s}$[bin] += [$h_i \mathbf{x}_i \mathbf{x}_i^T$, $g_i \mathbf{x}_i$]}
		\EndFor
	\end{algorithmic}
  \end{algorithm}
  \vspace{-4.5mm}	
  \begin{algorithm}[H]
    \caption{SIMD with Reduced Cache Misses}
    \begin{algorithmic}[1] 
		\State{\textbf{Input}: $\mathbf{x}_1,...,\mathbf{x}_n$, $g_1,...,g_n$, $h_1,...,h_n$, $leafBin_{s,j}$} 
		\State{\textbf{Output}: histogram $hist_{j,s}$}
		\For{i = 1 to n}
			\State{bin = $leafBin_{s,j}$[i]} 
			\State{$hist_{j,s}$[bin] += [$h_i \mathbf{x}_i \mathbf{x}_i^T$, $g_i \mathbf{x}_i$]  //SIMD add} 
		\EndFor
	\end{algorithmic}
  \end{algorithm}
\vspace{-4.5mm}	
\begin{figure}[h]
\centering
 	 \centering
 	 \includegraphics[width=0.8\linewidth]{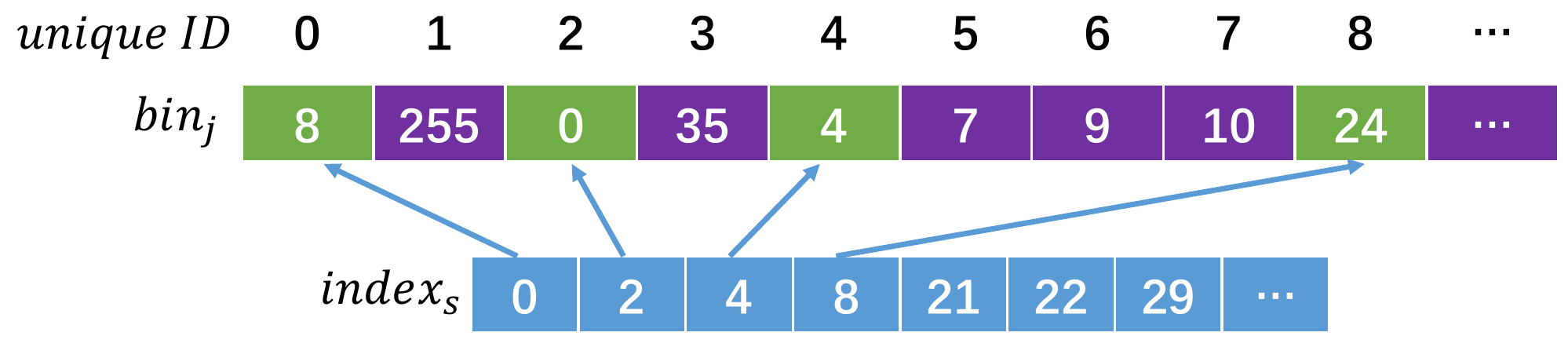}	
  	\captionof{figure}{Accessing $bin_j$ Causes Frequent Cache Misses}
 	 \label{binj}
\end{figure}
\vspace{-4.5mm} 
\begin{figure}[h]
 	 \centering
 	 \includegraphics[width=0.8\linewidth]{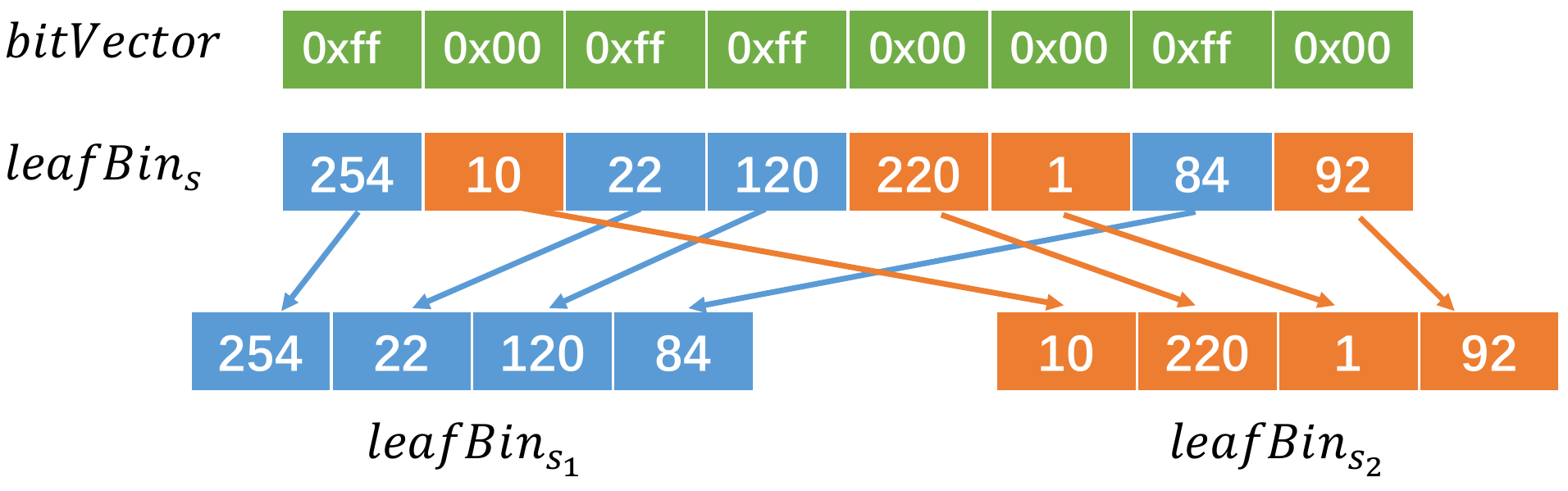}
  	\captionof{figure}{Parallel Bits Extract to Split $leafBin_{s,j}$} 
	\vspace{-2.5mm}
 	 \label{bmi}
\end{figure}
\subsection{SIMD Parallelism with Reduced Cache Misses}	
Single Instruction Multiple Data (SIMD) parallelism of modern CPUs supports operations on multiple data items with single instruction. 
It is obvious that SIMD can be used to speedup line 6 of Algorithm 2, which is a simultaneous addition of multiple items. With SIMD, however, each clock cycle more operands are needed, thus the speedup of SIMD is often bounded by memory bandwidth \cite{espasa1998vector}. In Algorithm 2, when accessing $bin_j$, we have to skip the data points not in leaf $s$ (purple blocks in Figure \ref{binj}). Thus the access to array $bin_j$ is discontinuous, causing frequent cache misses. To address this problem, we reduce cache misses by rearranging the data structures (a very similar idea is used in LightGBM \cite{ke2017lightgbm} for histogram construction of sparse features).

Suppose leaf $s$ has $n$ data points. For each feature $j$, we maintain an array $leafBin_{s,j}$ of length $n$ to record bin indices of feature $j$ for data points in leaf $s$. 
In other words, with the notations in Algorithm 2, for $i=1,...,n$, $leafBin_{s,j}$[i] = $bin_j$[$index_s$[i]]. 
Since each bin index is stored in a byte, and the access to $leafBin_{s,j}$ is continuous, we keep the memory footprint very small and reduces cache misses. Also, with $leafBin_{s,j}$, we can avoid accessing the unique ID array $index_s$. 
Histogram construction with $leafBin_{s,j}$ using SIMD is summarized in Algorithm 3. 

For root node $s_0$, $leafBin_{s_0,j}$ is exactly $bin_j$. When leaf $s$ is split into $s_1$ and $s_2$, $leafBin_{s,j}$ is split into $leafBin_{s_1,j}$ and $leafBin_{s_2,j}$ accordingly. The split operation has to be done for every feature $j$. In Section 5.2, we show how to reduce the cost of splitting $leafBin_{s,j}$ using Bit Manipulation Instruction Set.

\subsection{Using Bit Manipulation Instructions} 
Splitting $leafBin_{s,j}$ requires extracting bin indices from $leafBin_{s,j}$ and store into $leafBin_{s_1,j}$ and $leafBin_{s_2,j}$. To do this, we need to know for each data point in $s$, whether it goes to $s_1$ or $s_2$. This information is recorded in a bit vector. Specifically, if $s$ is split with condition $\mathbf{x}_{i,k} \le c$, then $bitVector$[i]= $\mathbf{x}_{i,k} \le c$. Creating $bitVector$ only requires a single sweep of $[\mathbf{x}_{1,k},...,\mathbf{x}_{n,k}]$. 
Then for each feature $j$, bin indices in $leafBin_{s,j}$ are extracted according to $bitVector$. 
BMI is an extension of x86 instructions to speedup bit operations. We use Parallel Bits Extract (PEXT) of BMI to extract the bin indices. PEXT takes two 64-bit registers $a$ and $b$ as operands. For each bit in $a$ whose value is $1$, the corresponding bit in $b$ is extracted and stored in the output register. Each bin index is stored in a single byte. Each PEXT instruction can handle 64 bits simultaneously, so we can process 8 bin indices in $leafBin_{s,j}$ simultaneously. 
The workflow of using PEXT is shown in Figure \ref{bmi}. We first broadcast each bit in $bitVector$ into a byte, thus 1 becomes 0xff and 0 becomes 0x00. Then, with a PEXT instruction, we can extract $leafBin_{s_1}$. Then we negate the bits in $bitVector$ and extract  $leafBin_{s_2}$ using another PEXT operation.  

\section{Experiments} 
Our experiments aim to answer the following questions
\textbf{1}. How the optimization techniques influence the accuracy and efficiency of boosted PL Trees.
\textbf{2}. How is our algorithm compared with state-of-the-art GBDT packages including LightGBM, XGBoost and CatBoost.  
We evaluate our algorithm on 10 public datasets.
We name our algorithm GBDT-PL. 
Our code, details of experiment setting and datasets is available at the github page. 
\footnote{
https://github.com/GBDT-PL/GBDT-PL.git
} 
\begin{figure}
    \centering
    \includegraphics[scale=0.25]{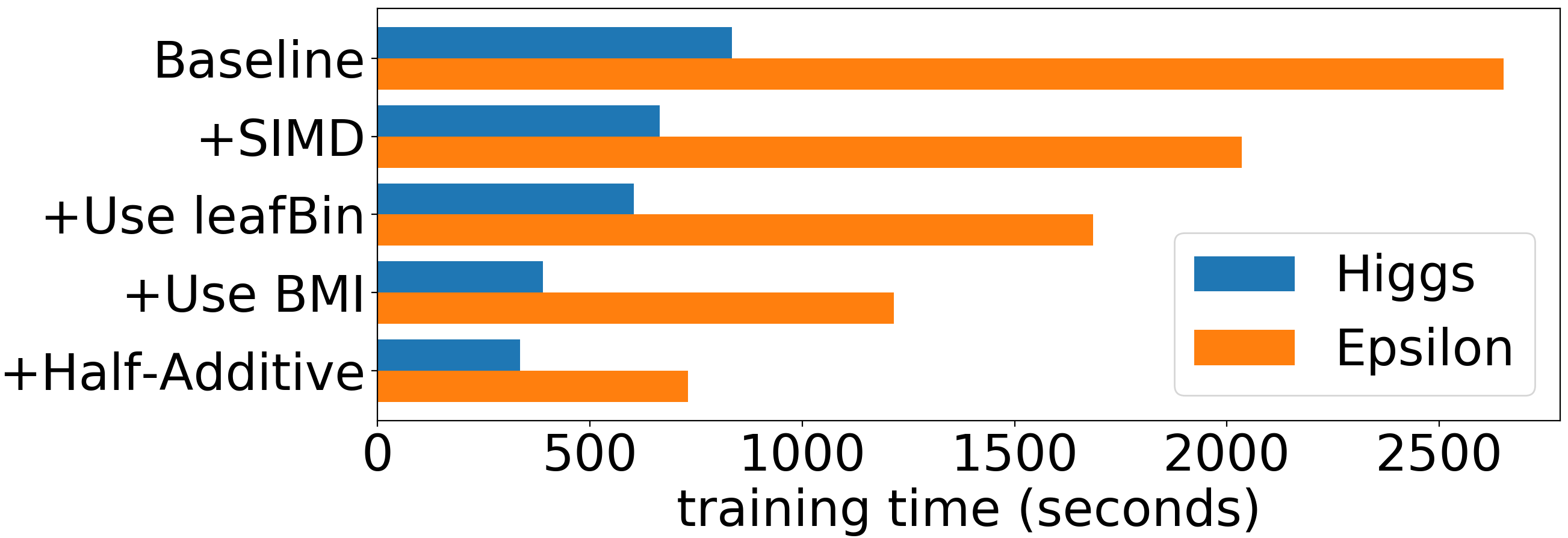} 
    \captionof{figure}{Speedup Effects of Optimization Techniques}
    \label{speedup}
\end{figure}
\subsection{Speedup Effects of Optimization Techniques}
To evaluate the speedup effects of various techniques in Section 4 and 5, we start from a baseline version, and add the optimization techniques incrementally. We record the training time, for 500 iterations using 63 histogram bins, of HIGGS and Epsilon datasets. Figure \ref{speedup} shows the training time when adding each optimization technique, from top to bottom. The first bar in the top is the baseline version (Algorithm 2 for histogram construction, using fully-corrective fitting). The second bar adds SIMD for histogram construction in Algorithm 2. The third bar uses the $leafBin$ (Algorithm 3). The fourth bar adds Bit Manipulation Instructions (BMI) (Section 5.2). The bottom bar adds the half-additive technique (Section 4.2). Compared with the second bar, the fourth bar with \textit{leafBin} data structure and BMI gets a speedup of {about 1.5 to 2} times. With \textit{leafBin} we reduce the cache misses when constructing
the histograms. With fewer cache misses when getting the bin indices, we can provide operands to the SIMD units in CPU more efficiently, thus further exploit the computation power.
And with BMI, we speedup the split operations of \textit{leafBin} thus reduces the overhead of maintaining \textit{leafBin}. 
\subsection{Effects of Optimization Techniques on Accuracy}
Incremental feature selection restricts the size of linear models, and half-additive fitting results in suboptimal linear model parameters. We evaluate the effects of these two techniques on the accuracy. We test the following 3 settings.
\textbf{a.} Disable the half-additive fitting and incremental feature selection. This means all features will be used in linear models of all nodes. 
\textbf{b.} Enable the incremental feature selection. Set the maximum constraint of regressors to 5.  
\textbf{c.} Based on \textbf{b}, enable the half-additive fitting.  
\begin{table}[t]
\small
\begin{tabular}{c|r|r|r}
\hline\hline
setting & time (s) & AUC & speedup\\
\hline
\textbf{a.} no feat. sel. and half-additive & 8443.75 & 0.859 & $\times 1.0$ \\
\hline
\textbf{b.} feat. sel., no half-additive & 574.88 & 0.856 & $\times 14.7$ \\
\hline
\textbf{c.} feat. sel., half-additive & 405.07 & 0.854 & $\times 20.8$ \\	
\hline \hline
\end{tabular}
\caption{Accuracy vs. Speedup}
\label{acc-speed}
\end{table}
Table \ref{acc-speed} shows the results. Here we use 255 histogram bins and 256 leaves. Incremental feature selection and half-additive fitting brings great speedup, with small sacrifice of accuracy. It is expensive to use all features in the linear models for leaves. With incremental feature selection, we restrict the size of linear models to reduce the computational cost.
With half-additive fitting, we fits linear models of any size with the cost of 3 regressors. These two techniques are important for the scalability of GBDT-PL. 
\begin{table*}[t!]
    \small
    \centering	
    \begin{tabular}{c|c|c|c|c|c|c|c|c|c|c}
\hline\hline 
Algorithm & Higgs & Hepmass & Casp & Epsilon  & Susy & CT & Sgemm & Year & SuperConductor & Energy\\
\hline
LightGBM & 0.854025 & 0.95563 & 3.4961 & 0.951422 & 0.878112 & 1.30902 & 4.61431 & 8.38817 & 8.80776 &  \textbf{64.256}\\
\hline
XGBoost & 0.854147 & 0.95567  & 3.4939 & 0.948292 & 0.877825 & 1.34131 & 4.37929 & 8.37935 & 8.91063 & 64.780 \\
\hline
CatBoost & 0.851590 & 0.95554 & 3.5183 & 0.957327 & 0.878206 & 1.36937 & 4.41177 & 8.42593 & \textbf{8.78452} & 65.761 \\
\hline
GBDT-PL & \textbf{0.860198} & \textbf{0.95652} & \textbf{3.4574} & \textbf{0.957894} & \textbf{0.878287} & \textbf{1.23753} & \textbf{4.16871} & \textbf{8.37233} & {8.79527} & 65.462 \\
\hline \hline
\end{tabular}
 \captionof{table}{Testing Accuracy}
\label{acc}
\end{table*} 
\begin{figure*}[ht!]
\centering	
  \subfigure[Higgs]{\includegraphics[scale=0.23]{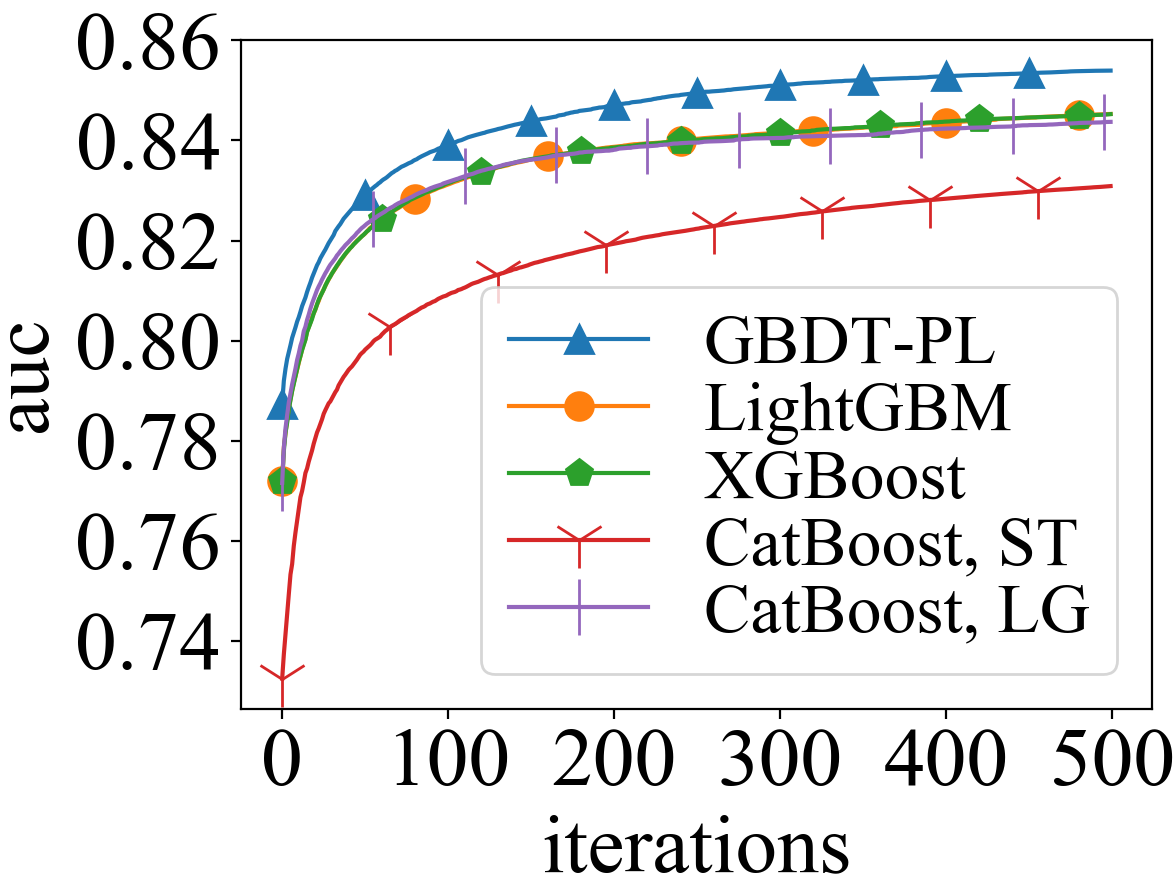}}\hspace{-0.5em}~
  \subfigure[Hepmass]{\includegraphics[scale=0.23]{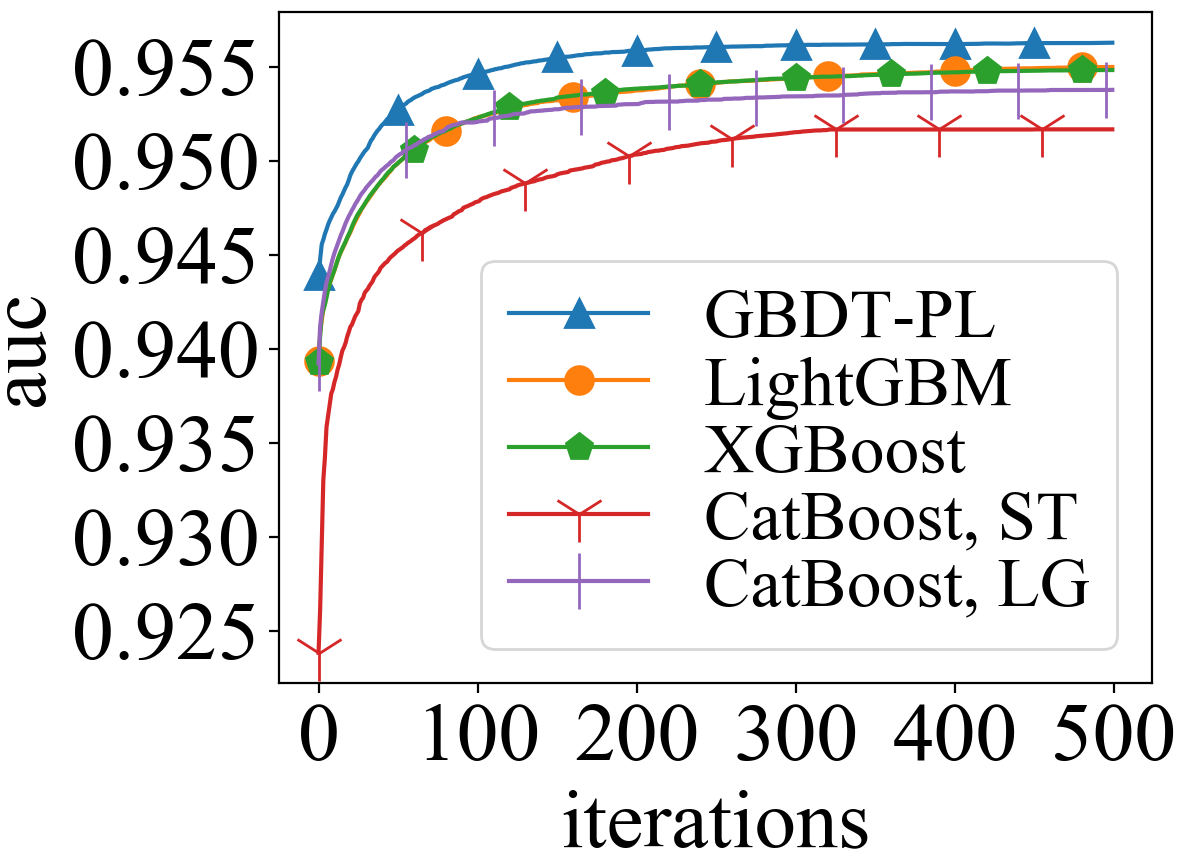}}\hspace{-0.5em}~
  \subfigure[Casp]{\includegraphics[scale=0.23]{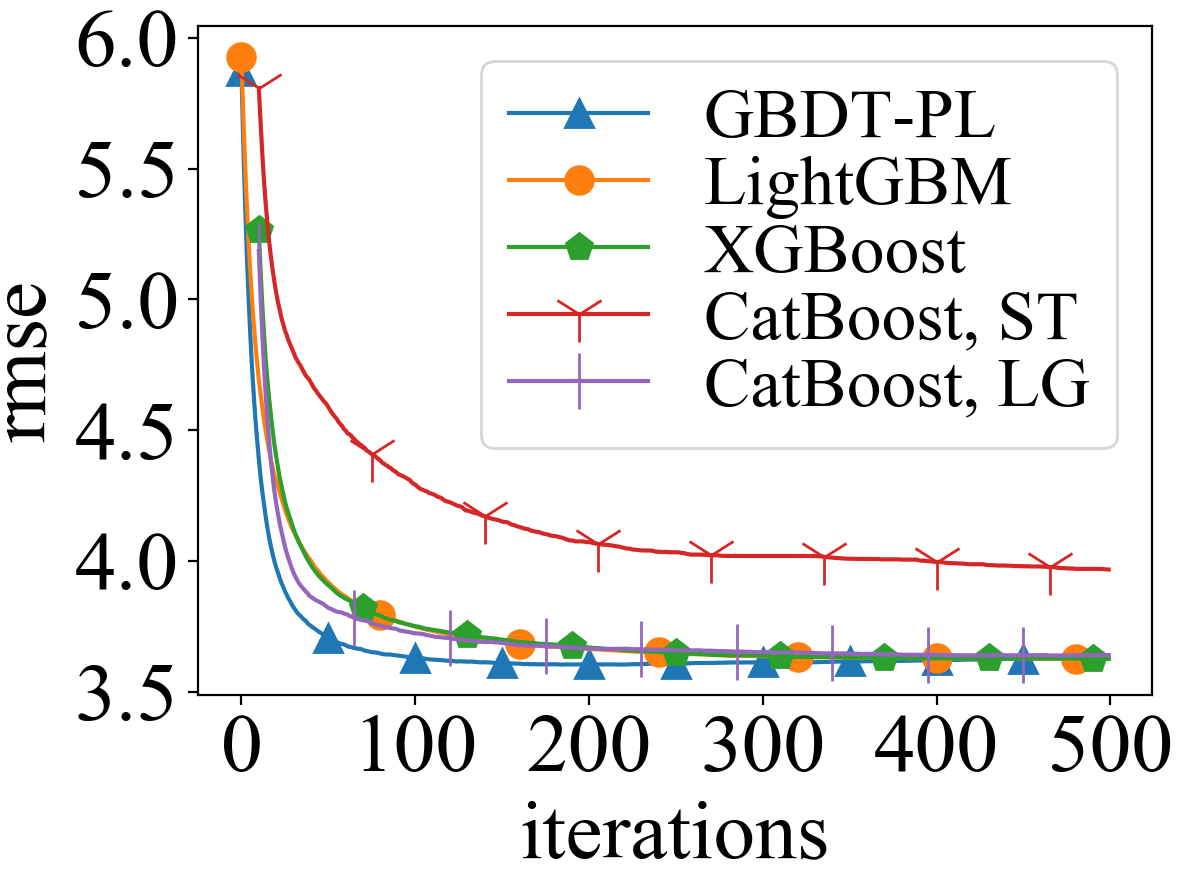}}\hspace{-0.5em}~
  \subfigure[Epsilon]{\includegraphics[scale=0.23]{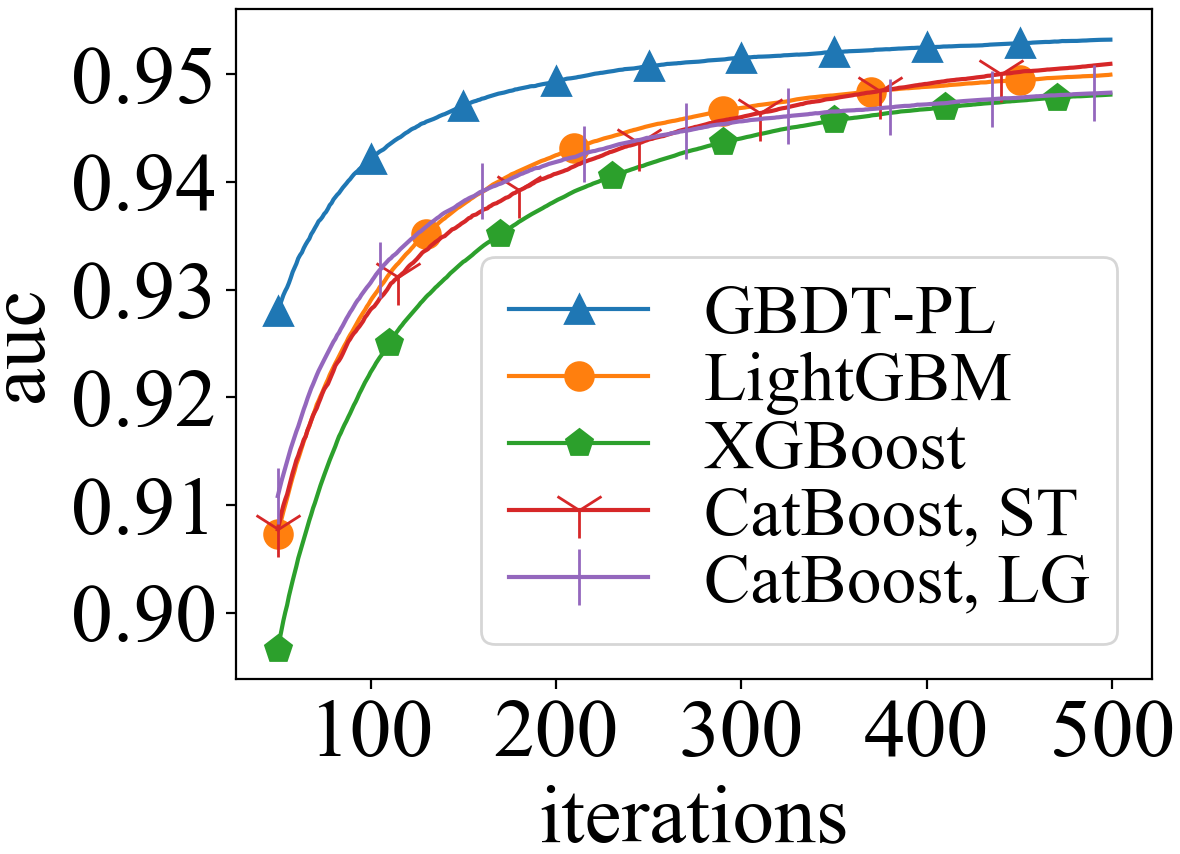}}\hspace{-0.5em}~
  \subfigure[Susy]{\includegraphics[scale=0.23]{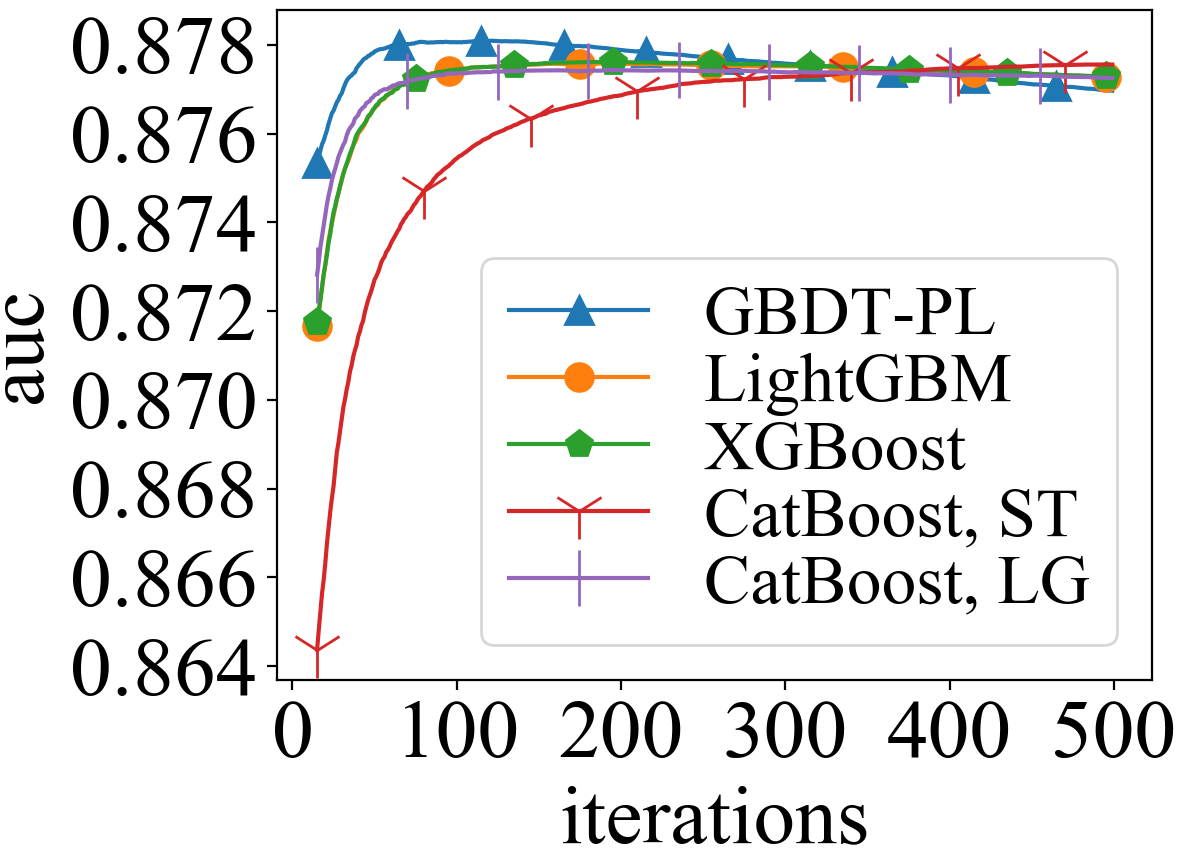}}\\	
  \subfigure[CT]{\includegraphics[scale=0.23]{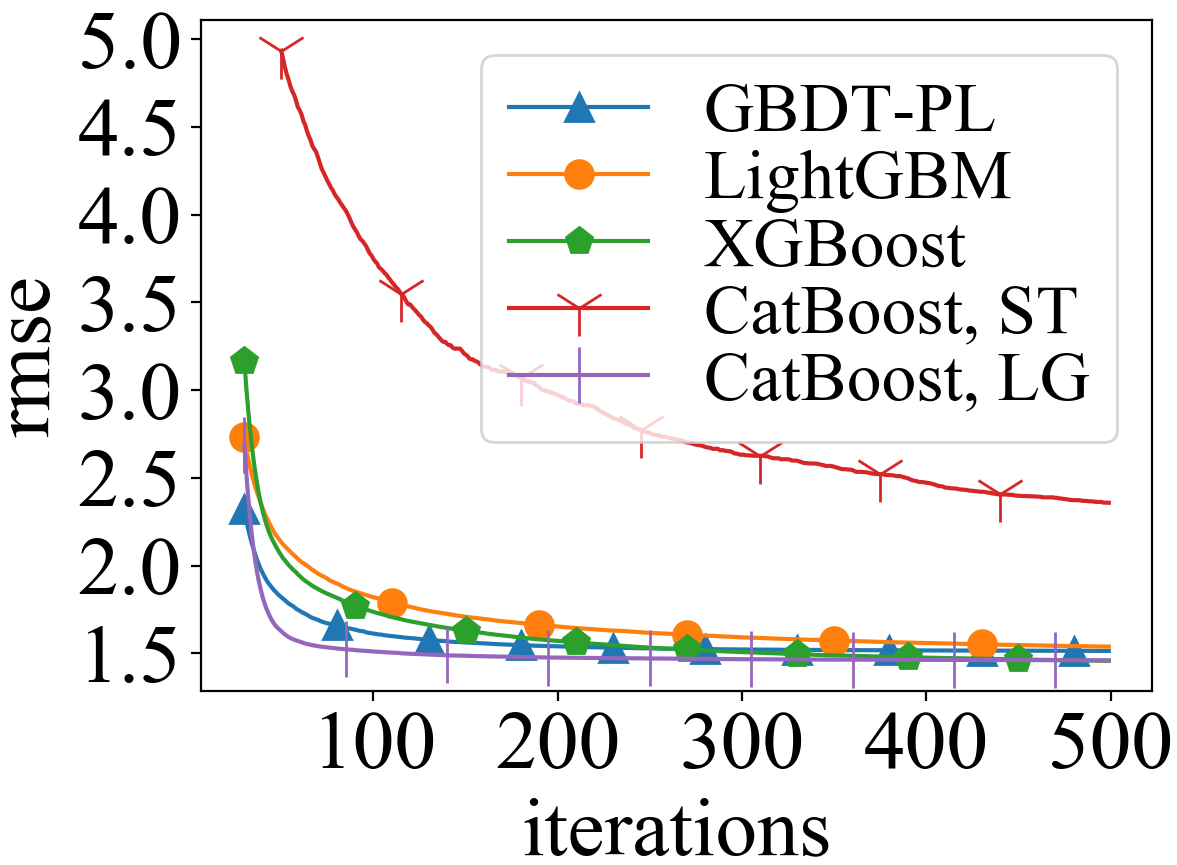}}\hspace{-0.5em}~
  \subfigure[Energy]{\includegraphics[scale=0.23]{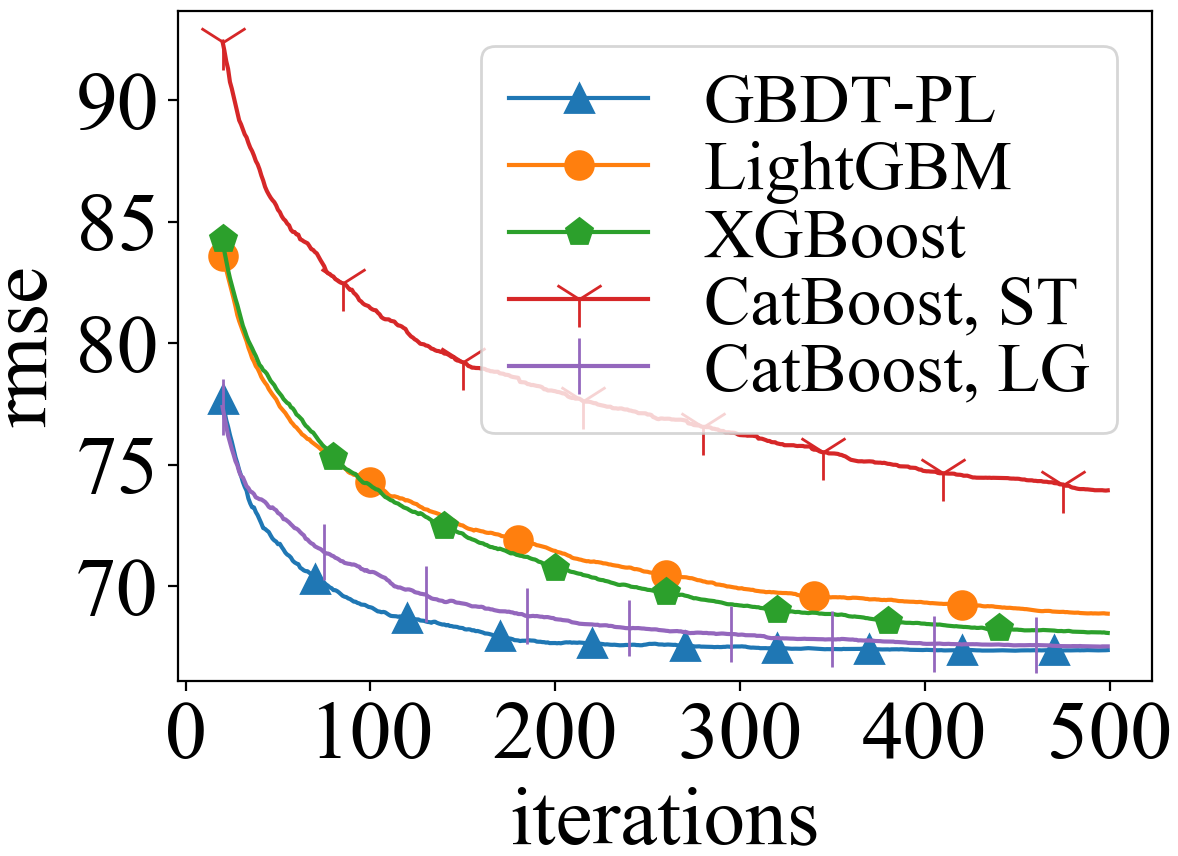}}\hspace{-0.5em}~
  \subfigure[SuperConductor]{\includegraphics[scale=0.23]{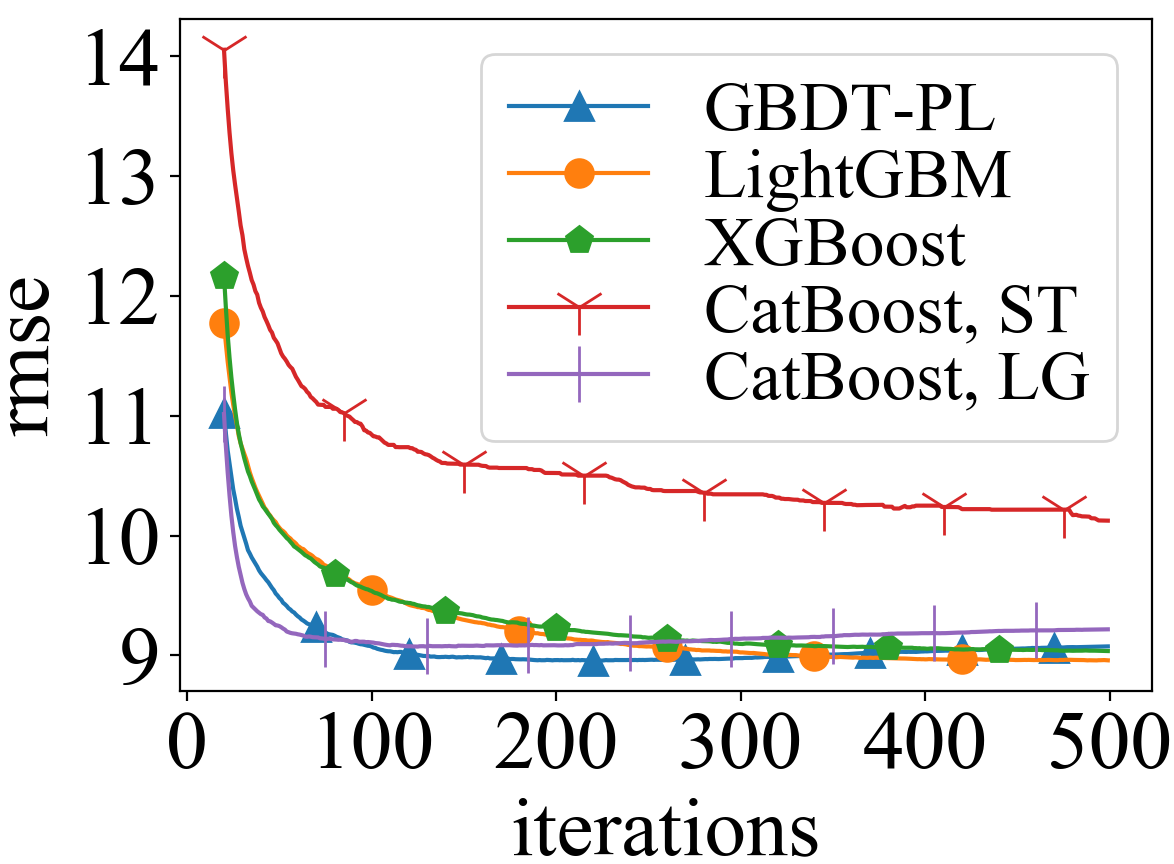}}\hspace{-0.5em}~
  \subfigure[Sgemm]{\includegraphics[scale=0.23]{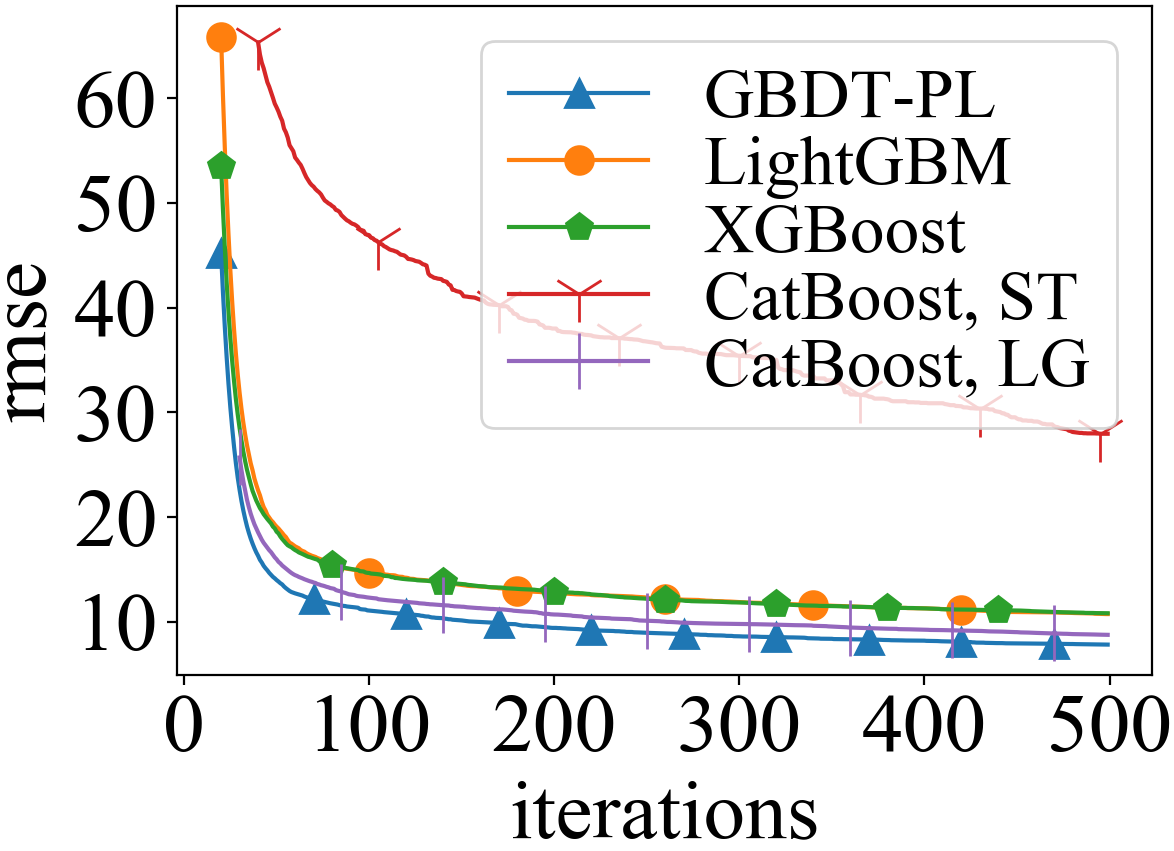}}\hspace{-0.5em}~
  \subfigure[Year]{\includegraphics[scale=0.23]{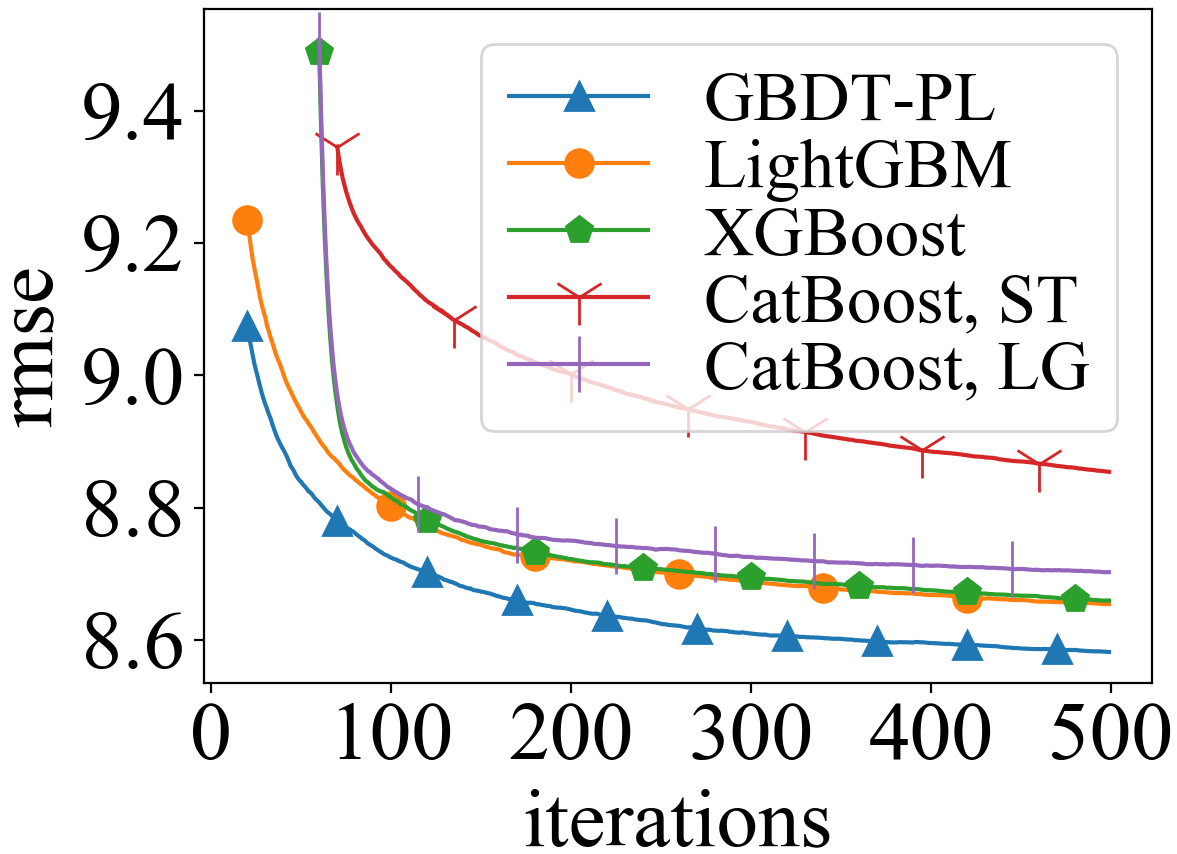}}\\
  \caption{Convergence Rate: AUC/RMSE per iteration (We use \textit{ST}	 for \textit{SymmetricTree} mode of CatBoost, and \textit{LG} for \textit{Lossguide} mode.)} 
  \label{conv} 
\end{figure*}
\begin{figure*}[ht!]
\centering	
\subfigure[Higgs]{\includegraphics[scale=0.23]{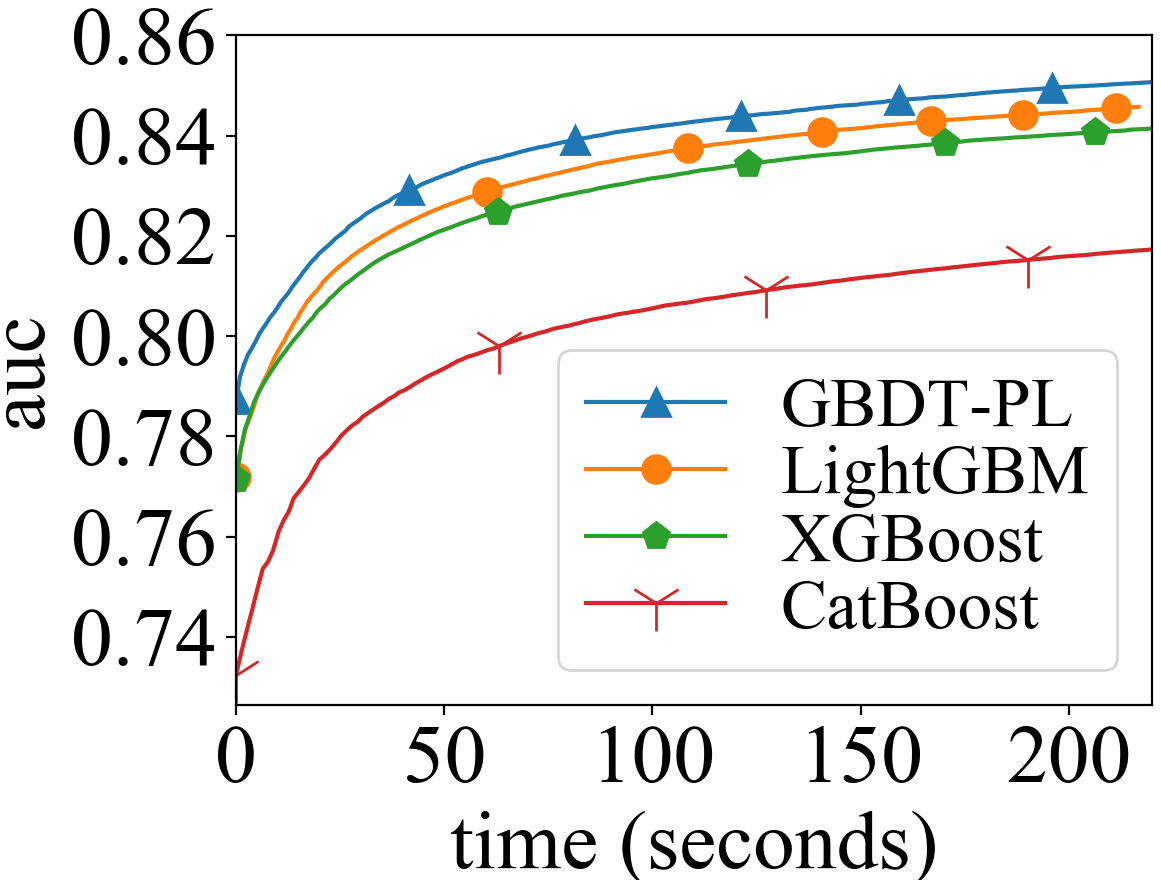}}\hspace{-0.2em}~
  \subfigure[Hepmass]{\includegraphics[scale=0.23]{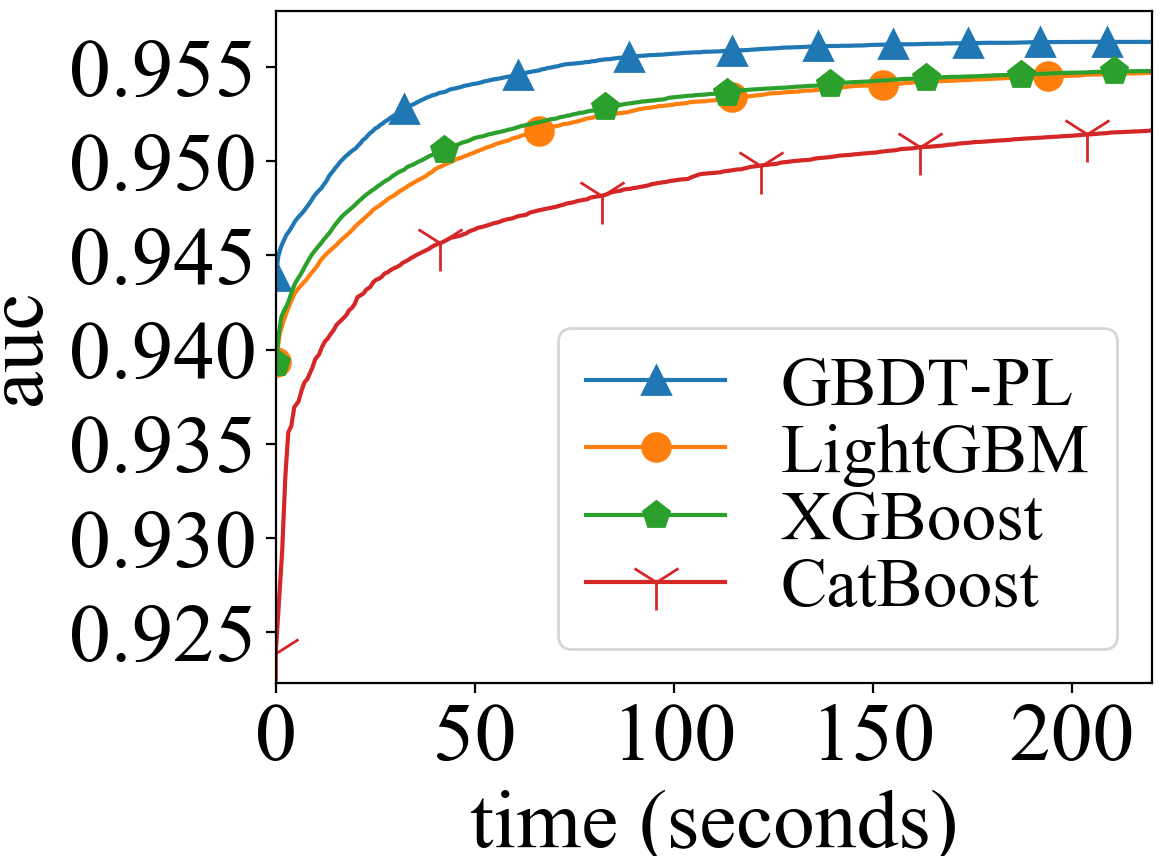}}\hspace{-0.2em}~
  \subfigure[Casp]{\includegraphics[scale=0.23]{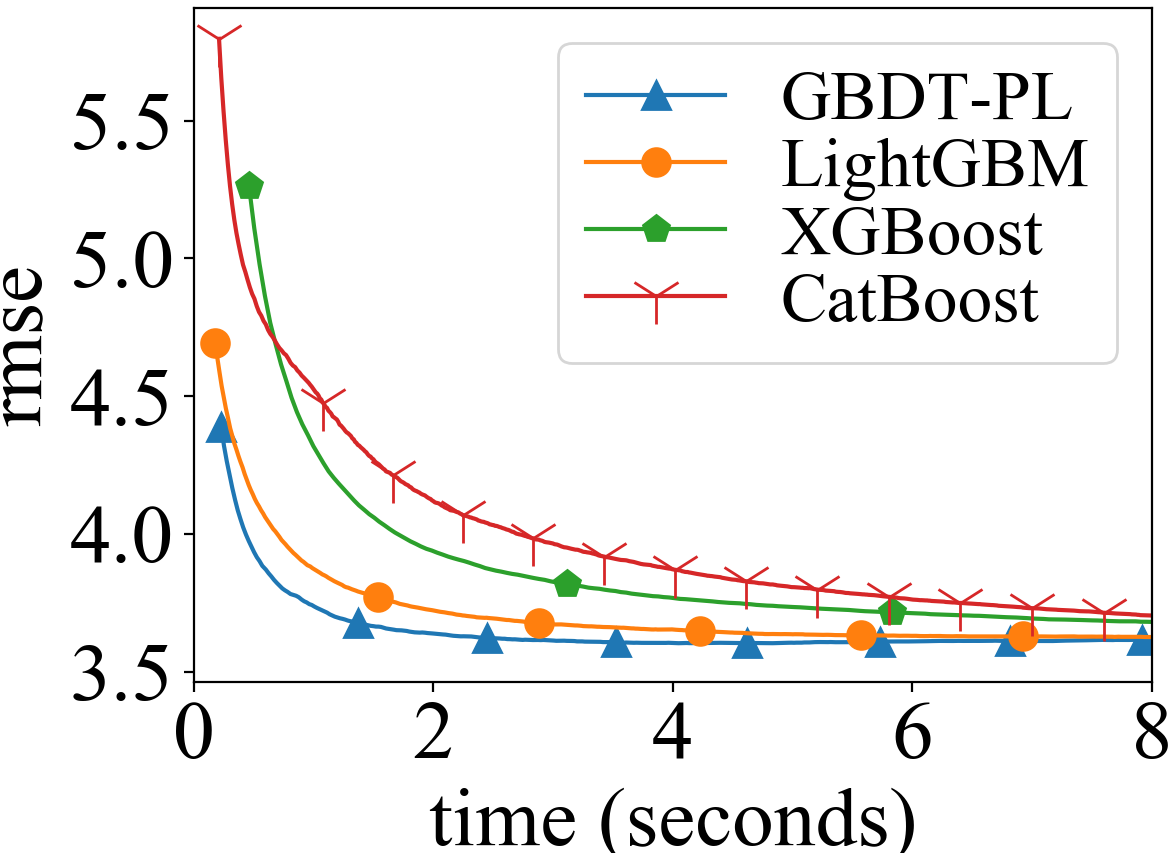}}\hspace{-0.2em}~
  \subfigure[Epsilon]{\includegraphics[scale=0.23]{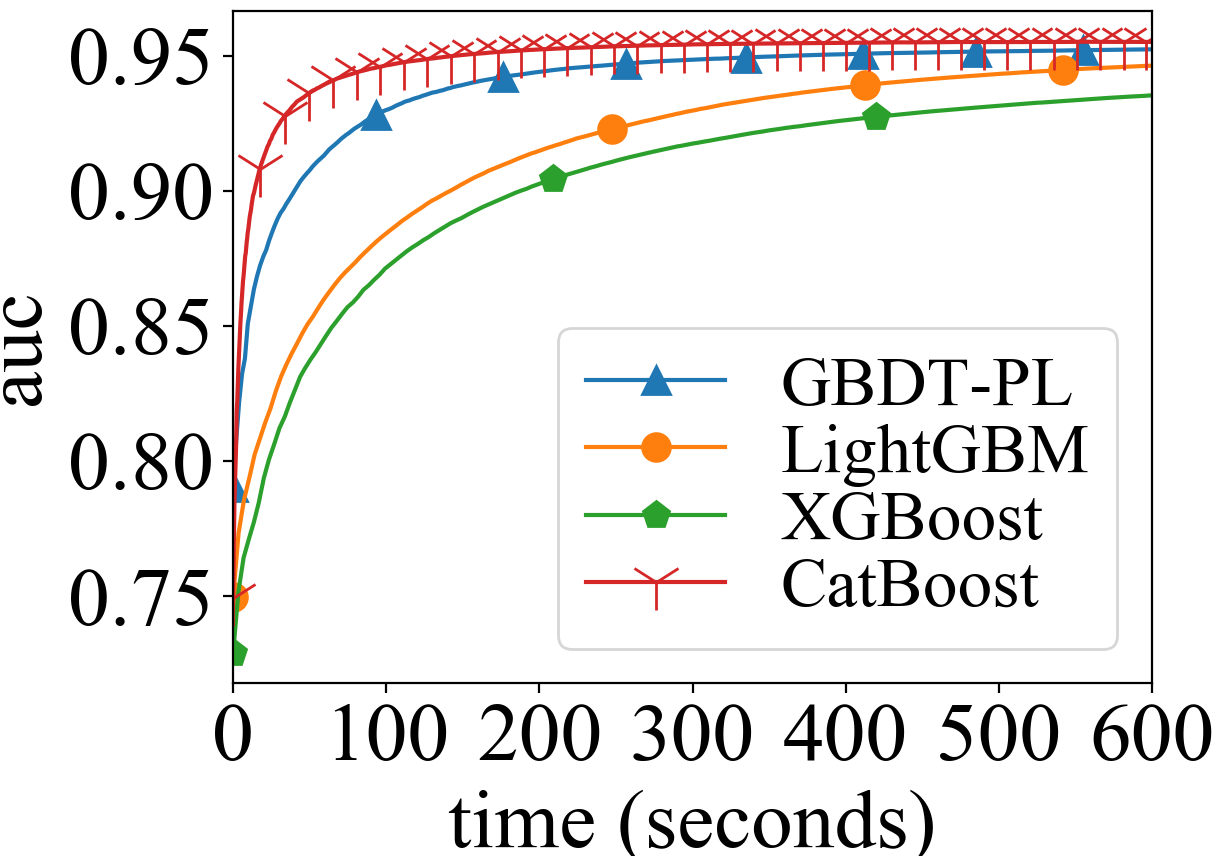}}\hspace{-0.2em}~
  \subfigure[Susy]{\includegraphics[scale=0.23]{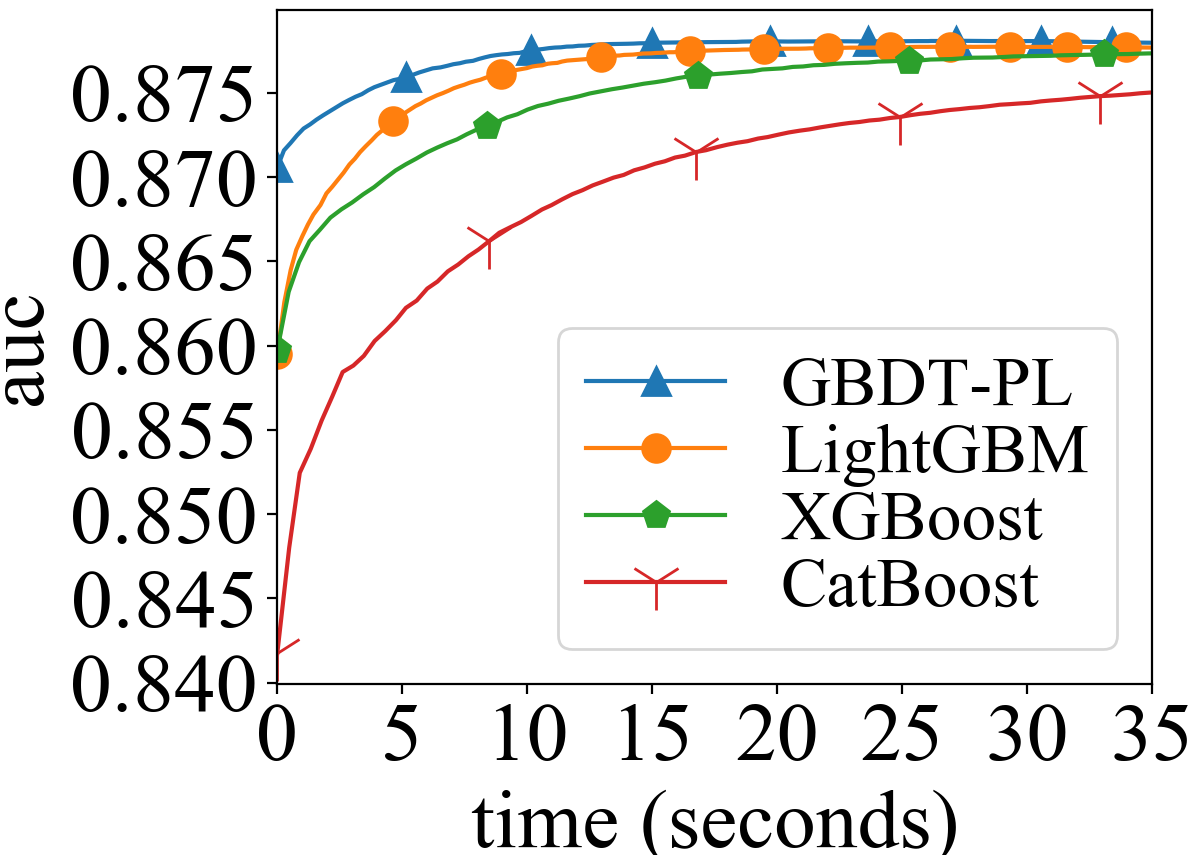}}\\	
  \subfigure[CT]{\includegraphics[scale=0.23]{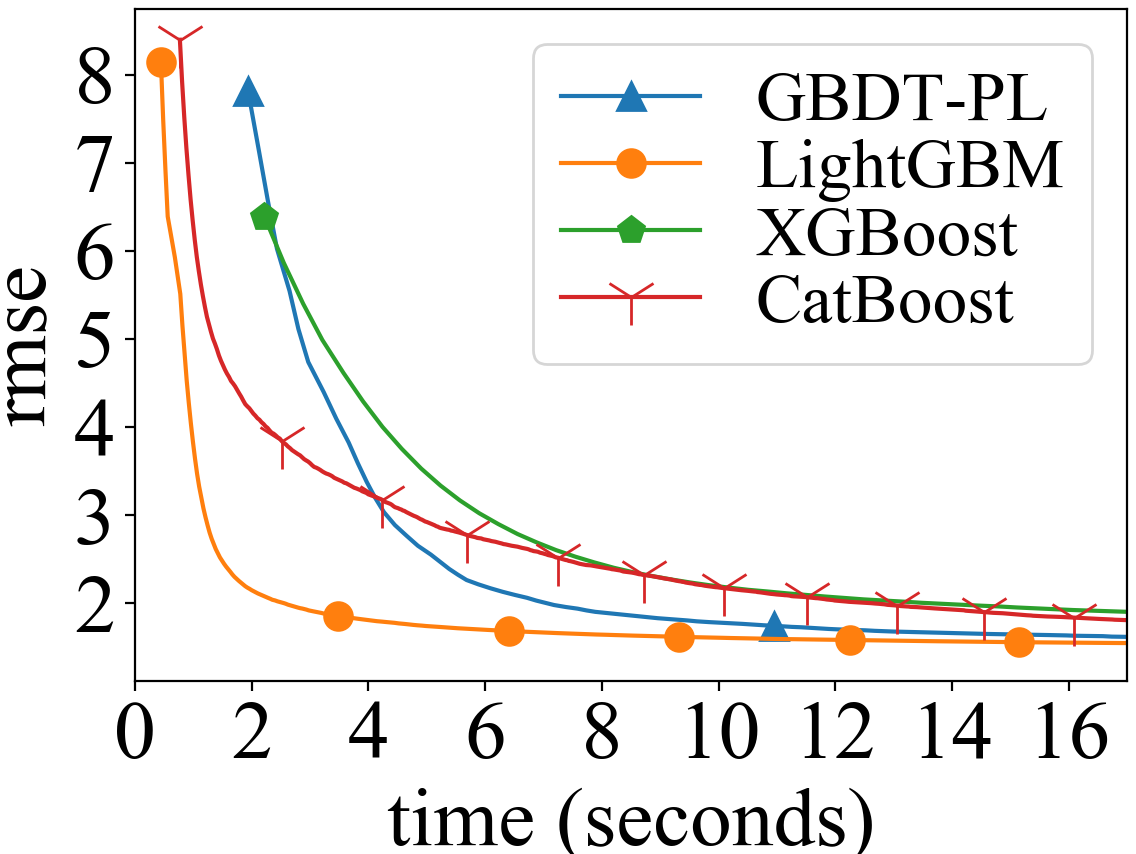}}\hspace{-0.2em}~
  \subfigure[Energy]{\includegraphics[scale=0.23]{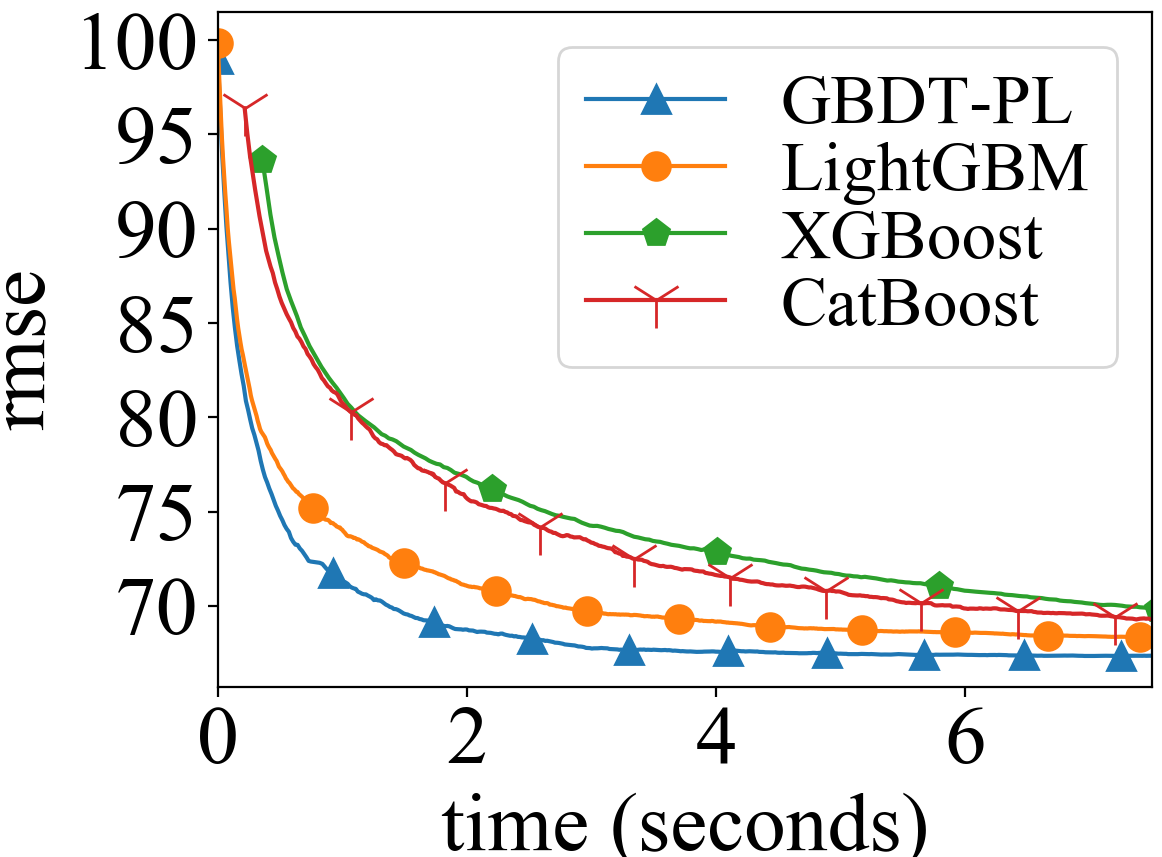}}\hspace{-0.2em}~
  \subfigure[SuperConductor]{\includegraphics[scale=0.23]{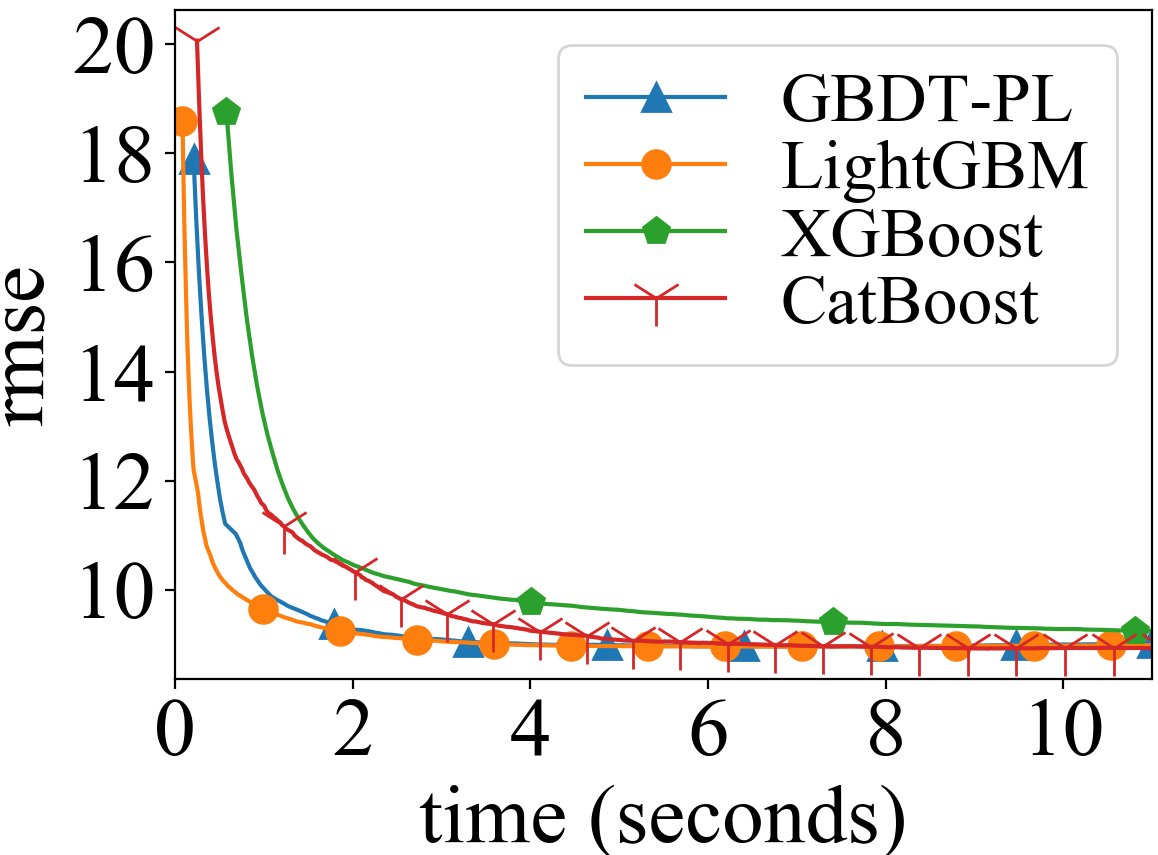}}\hspace{-0.2em}~	
  \subfigure[Sgemm]{\includegraphics[scale=0.23]{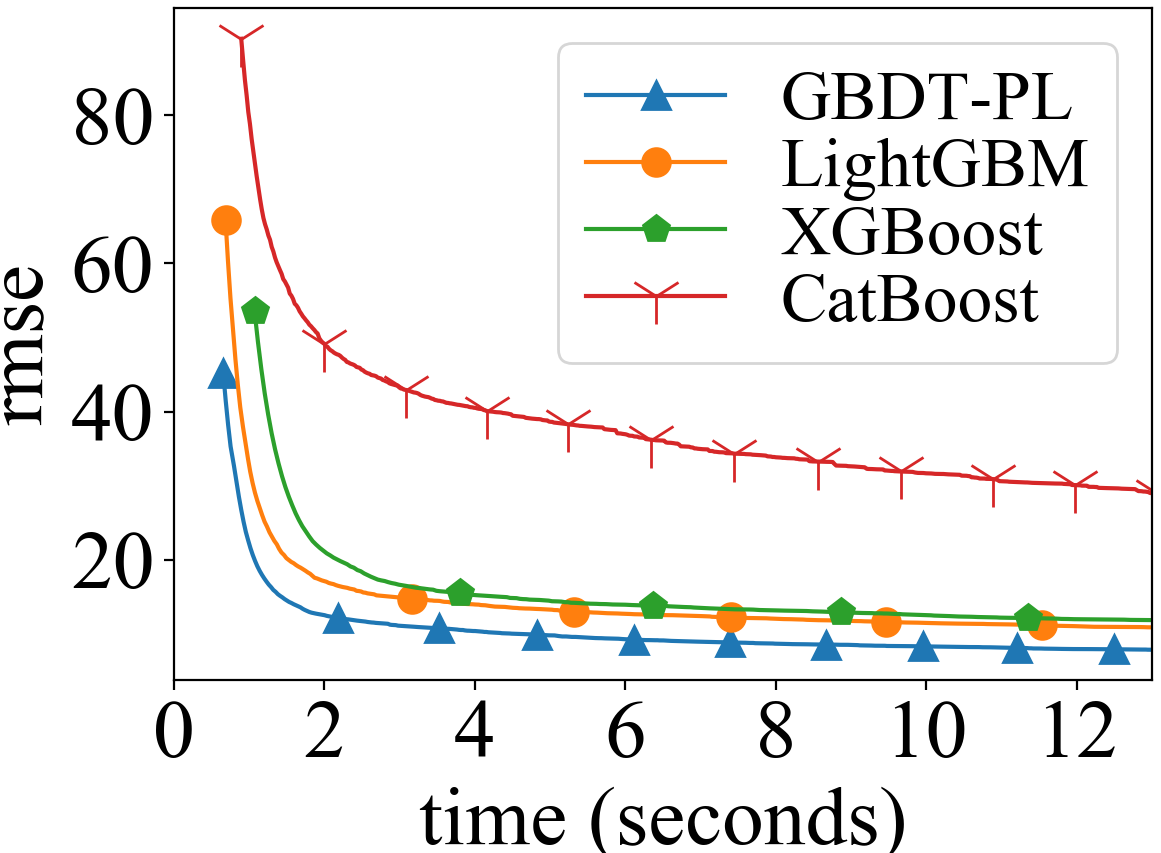}}\hspace{-0.2em}~
  \subfigure[Year]{\includegraphics[scale=0.23]{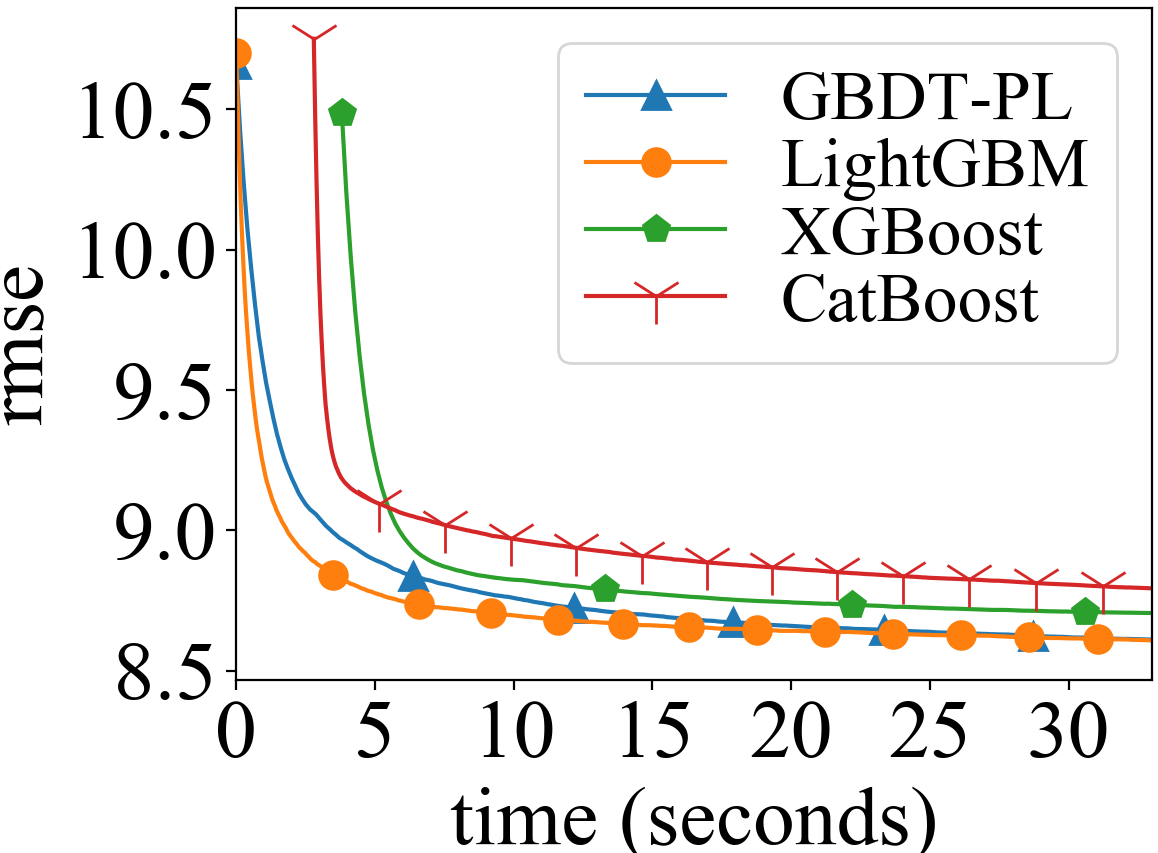}}\\
  \caption{Training Time Comparison on CPU: AUC/RMSE by training time}
  \label{res}
\end{figure*}
\subsection{Overall Performance}  
In this section, we compare GBDT-PL with XGBoost, LightGBM and CatBoost, which are state-of-the-art GBDT packages. 
We compare testing accuracy, convergence rate, and training speed on CPUs. 
\subsubsection{Accuracy} 	
We evaluate all 10 datasets with the 4 algorithms. For regression datasets Casp, CT, Sgemm, Year, SuperConductor and Energy, we use RMSE as evaluation metric. And for binary classification datasets Higgs, Hepmass, Epsilon and Susy, we use AUC.
Different settings of hyperparameters are tried. Key hyperparameters we tuned include:
1. $\textit{num leaves}\in \{16,64,256,1024\}$, which controls the size of each tree. For CatBoost with \textit{SymmetricTree} mode, the tree is grown by level, so $\textit{max depth}\in \{4,6,8,10\}$ is used instead of \textit{num leaves}. 2. $\textit{max bin}\in \{63,255\}$, the maximum number of bins in histograms. 3. $\textit{min sum hessians}\in \{1.0, 100.0\}$, the sum of hessians of data in each leaf.
4. $\textit{learning rate}\in \{0.01, 0.05, 0.1\}$, the weight of each tree. 5. $\textit{l2 reg}\in \{0.01, 10.0\}$, l2 regularization for leaf predicted values. 
We fix the number of regressors used in GBDT-PL to 5 in all runs. Different combinations of these parameters are tried. 
The maximum number of trees are chosen according to the rule $\textit{learning rate}\times \textit{num trees} = 50$ (For CatBoost in \textit{SymmetricTree} mode we use $\textit{learning rate}\times \textit{num trees} = 200$ since it converges slower). For large datasets, only learning rate 0.1 is tried. More details of parameter settings are listed in our github. 
For XGBoost, LightGBM and CatBoost, result of the best iteration over all settings on test set is recorded. For GBDT-PL, we seperate 20\% of training data for validation, and pick the best setting on validation set, then record the corresponding accuracy on test set. Table \ref{acc} shows the results. 
With linear models on leaves, GBDT-PL achieves better accuracy in these dense numerical datasets. It shows greater advantage in regression tasks. 
The results shows that 5 regressors is enough for most datasets. Adjusting the number of regressors can further improve the results. 
\subsubsection{Convergence Rate} 
To show that PL Trees speedup the convergence rate of GBDT, we set the maximum number of leaves to 256, and the maximum depth of CatBoost in \textit{SymmetricTree} mode to 8. We use 63 histogram bins, 500 iterations with learning rate 0.1. We set \textit{min sum hessians} to 100 and \textit{l2 reg} to 0.01. Figure 4 plots testing accuracy per iteration. In most datasets, GBDT-PL uses fewer trees to reach a comparable accuracy.
\subsubsection{Training Time on CPU} 
To test the efficiency on CPU, we use the same hyperparameters as previous subsection. (So far only \textit{SymmetricTree} mode is supported by CatBoost on CPU, so only this mode is tested.) Figure \ref{res} shows testing accuracy by training time.
With the same tree size, GBDT-PL achieves better accuarcy with less or comparable training time in most datasets. 
\section{Related Work and Discussions} 
Boosted PL Trees have several existing implementations. Weka \cite{hall2009weka} and Cubist \cite{kuhn2012cubist} use M5/M5' \cite{quinlan1992learning,wang1996induction} as base learners. M5/M5' grows the tree in a way similar to piece-wise constant ones, then fits linear models on the nodes after the tree structure has been fixed. By contrast, PL Tree used in our work is designed to greedily reduce the loss at each step of its growing. Whenever a node is split in our PL Tree, we choose the optimal split that will result in the largest reduction in loss, considering the linear models in both child nodes. \cite{wang2014boosted} proposes a gradient boosting algorithm using PL Trees as base learners and then apply it to a product demand prediction task. However, the algorithm only uses the first-order gradients. By contrast, GBDT-PL uses second-order gradients, which is important for faster convergence rate of GBDT \cite{sun2014convergence}. None of the aforementioned algorithms can handle large scale datasets as SOTA GBDT toolkits. Training scalable single PL Tree has been investigated in \cite{dobra2002secret}. To the best of our knowledge, GBDT-PL is the first to make boosted PL Trees scalable, by various optimization methods in Section 4 and 5. Currently GBDT-PL only handles numerical features. Both LightGBM and CatBoost handles categorical features. However, GBDT-PL is still feasible for categorical features by first converting them into numerical ones, as is done in CatBoost, which encodes the categorical values with label values.
Efficient training of GBDT on GPUs is also a hot topic \cite{zhang2017gpu,wen2018efficient}. We will add support for GPU in the future. 
\section{Conclusions} 
In this paper, we propose efficient GBDT with PL Trees. We first extend GBDT with second-order approximation for PL Trees. Then incremental feature selection and half-additive fitting are proposed to efficiently fit the linear models in tree nodes. Finally, we show how to exploit SIMD parallelism and reduce cache misses by rearranging the data structures with Bit Manipulation Instructions. The proposed optimization techniques are tested to show their effects on efficiency and accuracy. Comparisons with SOTA baselines show the value of our methods.  
\section*{Acknowledgements}
The research is supported in part by the National Basic Research Program of China Grant 2015CB358700, the National Natural Science Foundation of China Grant 61822203, 61772297, 61632016, 61761146003, and a grant from Microsoft Research Asia. 
\newpage
\bibliographystyle{named}
\bibliography{ijcai19}

\begin{thebibliography}{}

\bibitem[\protect\citeauthoryear{Breiman}{2017}]{breiman2017classification}
Leo Breiman.
\newblock {\em Classification and {Regression} {Trees}}.
\newblock Routledge, 2017.

\bibitem[\protect\citeauthoryear{Chen and Guestrin}{2016}]{chen2016xgboost}
Tianqi Chen and Carlos Guestrin.
\newblock {XGBoost}: A scalable tree boosting system.
\newblock In {\em Proceedings of the 22nd {ACM} {SIGKDD} International
  {Conference} on {Knowledge} {Discovery} and {Data} {Mining}}, pages 785--794.
  ACM, 2016.

\bibitem[\protect\citeauthoryear{Chen \bgroup \em et al.\egroup
  }{2012}]{chen2012combining}
Tianqi Chen, Linpeng Tang, Qin Liu, Diyi Yang, Saining Xie, Xuezhi Cao,
  Chunyang Wu, Enpeng Yao, Zhengyang Liu, Zhansheng Jiang, et~al.
\newblock Combining factorization model and additive forest for collaborative
  followee recommendation.
\newblock {\em KDD CUP}, 2012.

\bibitem[\protect\citeauthoryear{Dobra and Gehrke}{2002}]{dobra2002secret}
Alin Dobra and Johannes Gehrke.
\newblock Secret: a scalable linear regression tree algorithm.
\newblock In {\em Proceedings of the 8th ACM SIGKDD {International}
  {Conference} on {Knowledge} {Discovery} and {Data} {Mining}}, pages 481--487.
  ACM, 2002.

\bibitem[\protect\citeauthoryear{Espasa \bgroup \em et al.\egroup
  }{1998}]{espasa1998vector}
Roger Espasa, Mateo Valero, and James~E Smith.
\newblock Vector architectures: past, present and future.
\newblock In {\em Proceedings of the 12th {International} {Conference} on
  Supercomputing}, pages 425--432. ACM, 1998.

\bibitem[\protect\citeauthoryear{Friedman \bgroup \em et al.\egroup
  }{2000}]{friedman2000additive}
Jerome Friedman, Trevor Hastie, Robert Tibshirani, et~al.
\newblock Additive logistic regression: a statistical view of boosting (with
  discussion and a rejoinder by the authors).
\newblock {\em The {Annals} of {Statistics}}, 28(2):337--407, 2000.

\bibitem[\protect\citeauthoryear{Friedman}{1979}]{friedman1979tree}
Jerome~H Friedman.
\newblock A tree-structured approach to nonparametric multiple regression.
\newblock {\em Smoothing {Techniques} for {Curve} {Estimation}}, 757:5--22,
  1979.

\bibitem[\protect\citeauthoryear{Friedman}{2001}]{friedman2001greedy}
Jerome~H Friedman.
\newblock Greedy function approximation: a gradient boosting machine.
\newblock {\em Annals of {Statistics}}, pages 1189--1232, 2001.

\bibitem[\protect\citeauthoryear{Hall \bgroup \em et al.\egroup
  }{2009}]{hall2009weka}
Mark Hall, Eibe Frank, Geoffrey Holmes, Bernhard Pfahringer, Peter Reutemann,
  and Ian~H Witten.
\newblock The weka data mining software: an update.
\newblock {\em ACM SIGKDD {Explorations} {Newsletter}}, 11(1):10--18, 2009.

\bibitem[\protect\citeauthoryear{Ke \bgroup \em et al.\egroup
  }{2017}]{ke2017lightgbm}
Guolin Ke, Qi~Meng, Thomas Finley, Taifeng Wang, Wei Chen, Weidong Ma, Qiwei
  Ye, and Tie-Yan Liu.
\newblock Lightgbm: A highly efficient gradient boosting decision tree.
\newblock In {\em Advances in Neural Information Processing Systems}, pages
  3149--3157, 2017.

\bibitem[\protect\citeauthoryear{Kuhn \bgroup \em et al.\egroup
  }{2012}]{kuhn2012cubist}
Max Kuhn, Steve Weston, Chris Keefer, and Nathan Coulter.
\newblock Cubist models for regression.
\newblock {\em R package Vignette R package version 0.0}, 18, 2012.

\bibitem[\protect\citeauthoryear{Kuhn \bgroup \em et al.\egroup
  }{2018}]{kuhn2018package}
Max Kuhn, Steve Weston, Chris Keefer, and Maintainer~Max Kuhn.
\newblock Package ‘cubist’.
\newblock 2018.

\bibitem[\protect\citeauthoryear{Prokhorenkova \bgroup \em et al.\egroup
  }{2018}]{prokhorenkova2018catboost}
Liudmila Prokhorenkova, Gleb Gusev, Aleksandr Vorobev, Anna~Veronika Dorogush,
  and Andrey Gulin.
\newblock Catboost: unbiased boosting with categorical features.
\newblock In {\em Advances in Neural Information Processing Systems}, pages
  6639--6649, 2018.

\bibitem[\protect\citeauthoryear{Quinlan and
  others}{1992}]{quinlan1992learning}
John~R Quinlan et~al.
\newblock Learning with continuous classes.
\newblock In {\em 5th Australian {Joint} {Conference} on {Artificial}
  {Intelligence}}, volume~92, pages 343--348. World Scientific, 1992.

\bibitem[\protect\citeauthoryear{Quinlan}{1986}]{quinlan1986induction}
J.~Ross Quinlan.
\newblock Induction of decision trees.
\newblock {\em Machine {Learning}}, 1(1):81--106, 1986.

\bibitem[\protect\citeauthoryear{Quinlan}{2014}]{quinlan2014c4}
J~Ross Quinlan.
\newblock {\em C4. 5: programs for machine learning}.
\newblock Elsevier, 2014.

\bibitem[\protect\citeauthoryear{Sun \bgroup \em et al.\egroup
  }{2014}]{sun2014convergence}
Peng Sun, Tong Zhang, and Jie Zhou.
\newblock A convergence rate analysis for logitboost, mart and their variant.
\newblock In {\em ICML}, pages 1251--1259, 2014.

\bibitem[\protect\citeauthoryear{Tyree \bgroup \em et al.\egroup
  }{2011}]{tyree2011parallel}
Stephen Tyree, Kilian~Q Weinberger, Kunal Agrawal, and Jennifer Paykin.
\newblock Parallel boosted regression trees for web search ranking.
\newblock In {\em Proceedings of the 20th {International} {Conference} on World
  {Wide} {Web}}, pages 387--396. ACM, 2011.

\bibitem[\protect\citeauthoryear{Vens and Blockeel}{2006}]{vens2006simple}
Celine Vens and Hendrik Blockeel.
\newblock A simple regression based heuristic for learning model trees.
\newblock {\em Intelligent Data Analysis}, 10(3):215--236, 2006.

\bibitem[\protect\citeauthoryear{Vogel \bgroup \em et al.\egroup
  }{2007}]{vogel2007scalable}
David~S Vogel, Ognian Asparouhov, and Tobias Scheffer.
\newblock Scalable look-ahead linear regression trees.
\newblock In {\em Proceedings of the 13th ACM SIGKDD {International}
  {Conference} on Knowledge {Discovery} and {Data} {Mining}}, pages 757--764.
  ACM, 2007.

\bibitem[\protect\citeauthoryear{Wang and Hastie}{2014}]{wang2014boosted}
Jianqiang~C Wang and Trevor Hastie.
\newblock Boosted varying-coefficient regression models for product demand
  prediction.
\newblock {\em Journal of Computational and Graphical Statistics},
  23(2):361--382, 2014.

\bibitem[\protect\citeauthoryear{Wang and Witten}{1997}]{wang1996induction}
Y.~Wang and I.~H. Witten.
\newblock Induction of model trees for predicting continuous classes.
\newblock In {\em Poster papers of the 9th {European} {Conference} on {Machine}
  {Learning}}. Springer, 1997.

\bibitem[\protect\citeauthoryear{Wang \bgroup \em et al.\egroup
  }{2014}]{wang2014intel}
Endong Wang, Qing Zhang, Bo~Shen, Guangyong Zhang, Xiaowei Lu, Qing Wu, and
  Yajuan Wang.
\newblock Intel math kernel library.
\newblock In {\em High-Performance Computing on the Intel{\textregistered} Xeon
  Phi™}, pages 167--188. Springer, 2014.

\bibitem[\protect\citeauthoryear{Wen \bgroup \em et al.\egroup
  }{2018}]{wen2018efficient}
Zeyi Wen, Bingsheng He, Ramamohanarao Kotagiri, Shengliang Lu, and Jiashuai
  Shi.
\newblock Efficient gradient boosted decision tree training on {GPUs}.
\newblock In {\em 2018 IEEE International Parallel and Distributed Processing
  Symposium (IPDPS)}, pages 234--243. IEEE, 2018.

\bibitem[\protect\citeauthoryear{Zhang \bgroup \em et al.\egroup
  }{2018}]{zhang2017gpu}
Huan Zhang, Si~Si, and Cho-Jui Hsieh.
\newblock {GPU}-acceleration for large-scale tree boosting.
\newblock In {\em SysML Conference}, 2018.

\end{thebibliography}
\newpage
\section*{Appendix}
\section*{A. More on Histograms for GBDT With PL Tree} 
Histogram is used in several GBDT implementations \cite{tyree2011parallel,chen2016xgboost,ke2017lightgbm} to reduce the number of potential split points. For each feature, a histogram of all its values in all data points is constructed. The boundaries of bins in the histogram are chosen to distribute the training data evenly over all bins. Each bin accumulates the statistics needed to calculate the loss reduction. When finding the optimal split point, we only consider the bin boundaries, instead of all unique feature values. After the histogram is constructed, we only need to record the bin number of each feature for each data point. Fewer than 256 bins in a histogram is enough to achieve good accuracy  \cite{zhang2017gpu}, thus a bin number can be stored in a single byte. We can discard the original feature values and only store the bin numbers during the boosting process. Thus using histograms produces small memory footprint. 

It seems nature to directly use the histogram technique in our algorithm. For LightGBM and XGBoost, each bin in a histogram only needs to record the sum of gradients and hessians of data in that bin. For our algorithm, the statistics in the histogram is more complex. The statistics are used to compute the least squares. Each bin $B$ needs to record both $\sum_{i\in B} h_i \mathbf{x}_i \mathbf{x}_i^T$ and $\sum_{i\in B} g_i \mathbf{x}_i$, where $\mathbf{x}_i$ is the column vector of selected regressors of data $i$. 
However, the feature values $\mathbf{x}_i$ are needed when fitting linear models in leaves. We still need to access the feature values constantly, which incurs long memory footprint. To overcome this problem, for each feature $j$ and each bin $i$ of $j$, we record the average feature values in bin $j$, denoted as $\overline{x}_{i,j}$. When fitting linear models, we use $\overline{x}_{i,j}$ to replace the original feature value $\mathbf{x}_{k,i}$. Here $\mathbf{x}_{k,i}$ is the value of feature $i$ of data point $\mathbf{x}_k$, and $\mathbf{x}_{k,i}$ falls in bin $j$. In this way, we can still discard the original feature values after preprocessing. Thus we adapt the histogram technique to PL Trees and preserve the small memory footprint. 

The histogram technique used in several existing methods (such as XGBoost and LightGBM) only record 2 elements in each bin (sum of gradients and hessians). Our histograms require more than 2 elements (including qudratic terms of feature values). So the detailed implementation is in fact quite different from existing ones. To better utilized the SIMD units when constructing histograms, we use intel intrinsics (which directly indicate the assemble instructions to use) to carefully arrange the calculation.

\section*{B. Experiment Platform for Training Time Recording}
The experiment environment for training time comparison are listed in Table 1. 
\begin{table}[h]
\scriptsize
\caption{Experiment Platform}
\centering
\begin{tabular}{c|c|c}
\hline\hline
OS & CPU & Memory \\
\hline
CentOS Linux 7 & 2 $\times$ Xeon E5-2690 v3 & DDR4 2400Mhz, 128GB\\
\hline\hline
\end{tabular}
\label{plat}
\end{table}  

\section*{C. Datasets}
The datasets we used in this paper are all from UCI datasets. The number of instances and features can be found in the following table. We will provide details about how we split the datasets into train and testing sets on our github page. \footnote{
https://github.com/GBDT-PL/GBDT-PL.git
} 
\begin{table}[h]
\scriptsize
\caption{Datasets Description} 
\centering
\begin{tabular}{c|c|c|c|c}
\hline\hline
name & \# training & \# testing & \# features & task\\
\hline
HIGGS & 10000000 & 500000 & 28 & classification \\
\hline
HEPMASS & 7000000 & 3500000 & 28 & classification \\
\hline
CASP & 30000 & 15731 & 9 & regression \\
\hline
Epsilon & 400000 & 100000 & 2000 & classification \\
\hline
SUSY & 4000000 & 1000000 & 18 & classification \\
\hline
SGEMM & 193280 & 48320 & 14 & regression \\
\hline
SUPERCONDUCTOR & 17008 & 4255 & 81 & regression \\
\hline
CT & 42941 & 10559 & 384 & regression \\	
\hline
Energy & 15788 & 3947 & 27 & regression \\
\hline
Year & 412206 & 103139 & 90 & regression \\
\hline\hline
\end{tabular}	
\label{data}
\end{table}
For datasets with features of large values, including Year, SUPERCONDUCTOR, Energy and CASP, we first find the minimum and maximum values of each feature in the training set, and rescale the features into range [0, 1] before feeding it into GBDT-PL.  This is for numerical stability when computing matrix inversions. 
\section*{D. Comparison With Existing Boosted PL Trees} 
We compare our results with boosted PL Trees in Weka \cite{hall2009weka} and Cubist \cite{kuhn2018package} packages. The  base learner of Weka and Cubist is M5/M5P, a PL Tree proposed by \cite{quinlan1992learning,wang1996induction}. The main differences between M5/M5P and our algorithm are: \textbf{1}. M5/M5P does not use half-additive fitting, histogram and our system optimization techniques. \textbf{2}. M5/M5P grows the tree in a way similar to piecewise constant regression trees (e.g. CART), then fits the linear models at the nodes. Each split in our algorithm considers how much the resultant linear models in the child nodes will reduce the boosting objective. In other words, the split finding in GBDT-PL is greedy and more expensive. 
\begin{figure}[h]	
  \subfigure[HIGGS-100k AUC]{\includegraphics[scale=0.25]{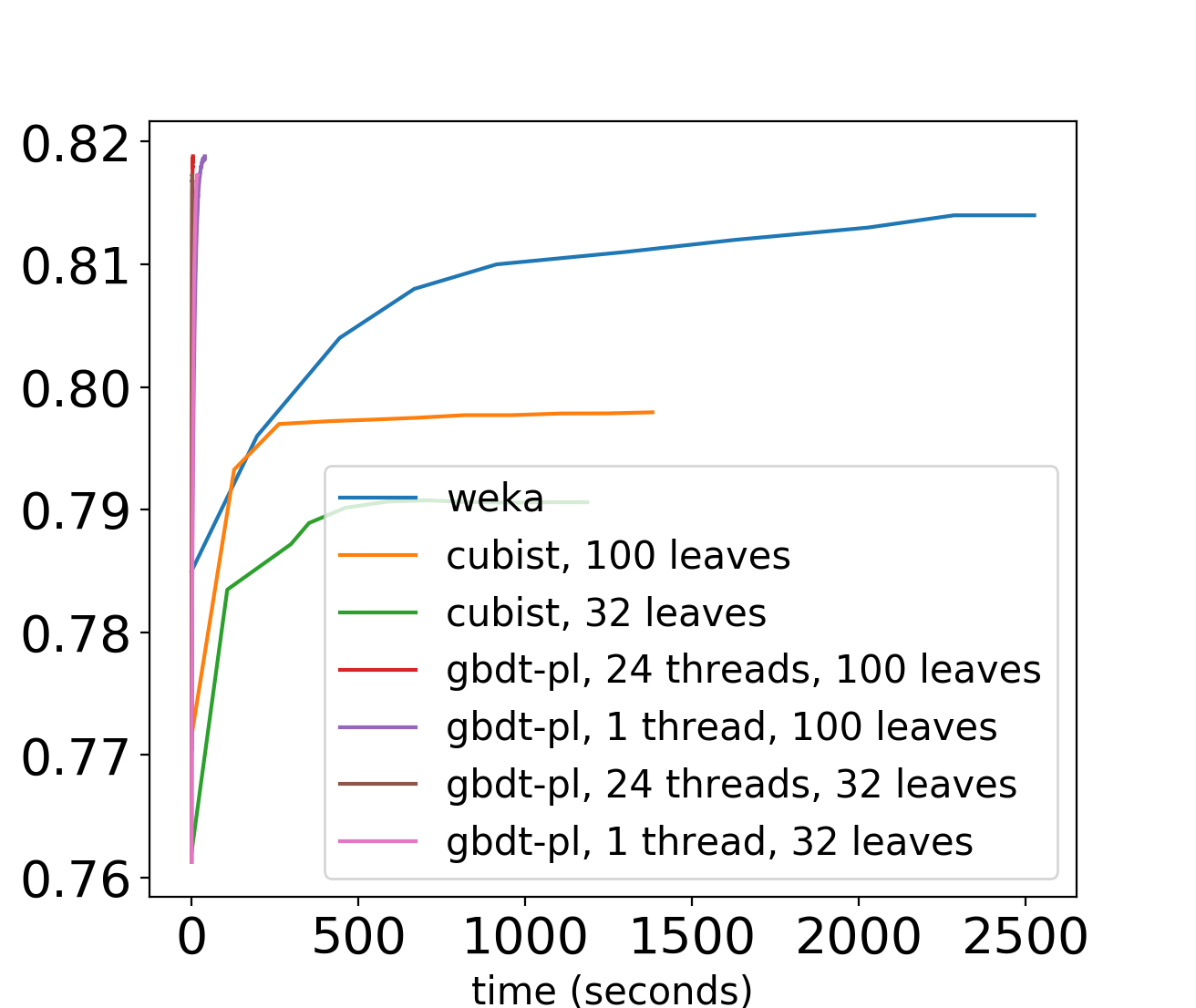}}~
  \subfigure[CASP RMSE]{\includegraphics[scale=0.25]{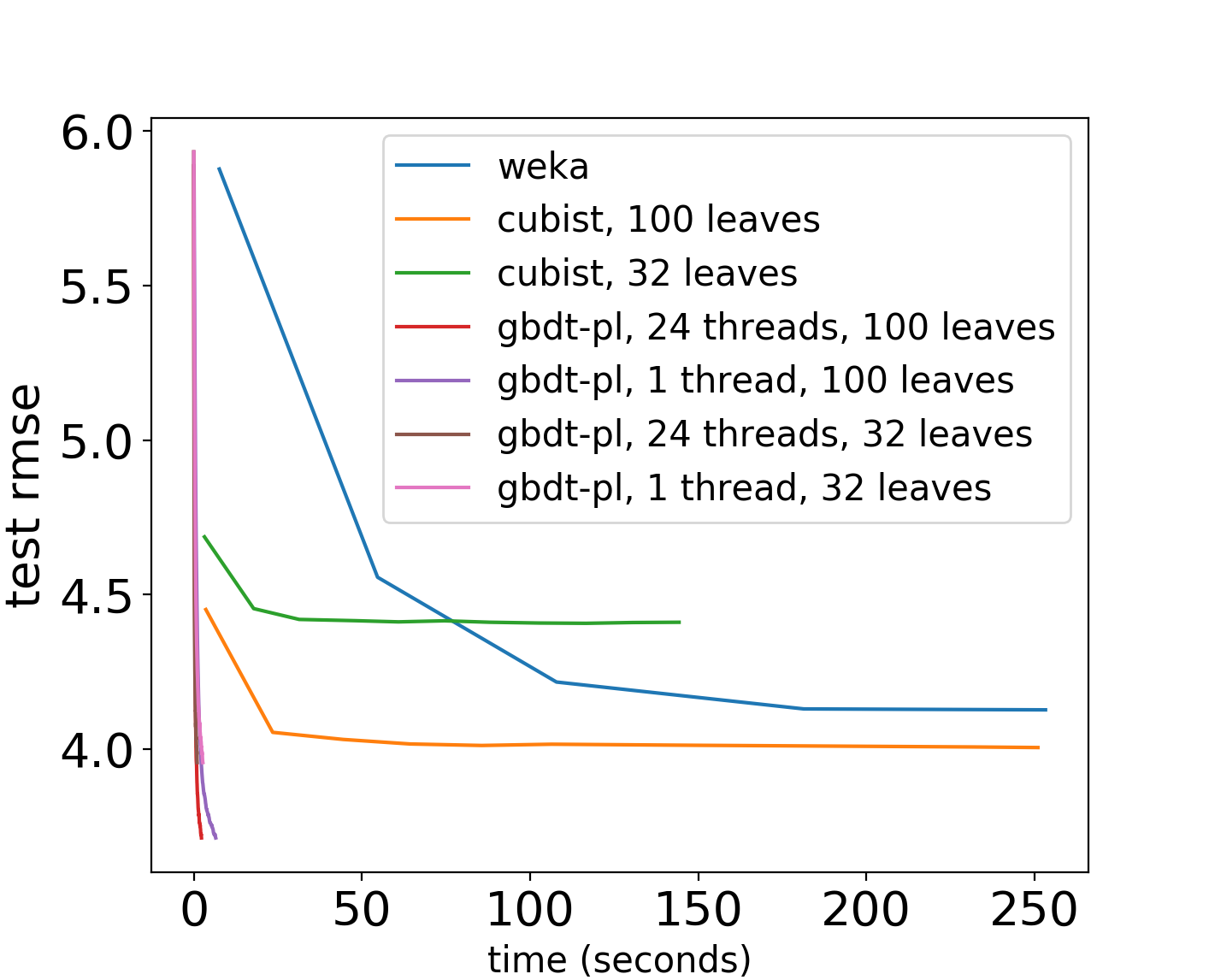}}\\
  \caption{Comparison With Weka and Cubist}	
  \label{weka}
\end{figure}
Figure \ref{weka} shows the results on a small subset of HIGGS dataset with only 100k samples and CASP. All algorithms train 100 trees. GBDT-PL has a significant advantage in efficiency. For example, GBDT-PL finished training 100 32-leaf trees for CASP within 2 seconds, while Cubist takes about 150 seconds. 
\section*{E. Parameter Settings}  
The parameters we tried are listed in following tables. For big datastes (HIGGS, Epsilon, SUSY and HEPMASS), we only test the learning rate 0.1. 
We evaluate all different combinations of these values. On small datasets, 96 combinations of parameters are tested for GBDT-PL,
and 144 combinations of parameters are tested for XGBoost, LightGBM and CatBoost. The total number of trees (iterations) are chosen according to the learning rate. 
For XGBoost, LightGBM and GBDT-PL, 500 trees for \textit{learning\_rate} 0.1, 1000 trees for \textit{learning\_rate} 0.05 and 5000 trees for \textit{learning\_rate} 0.01. 
For CatBoost in \textit{SymmetricTree} mode, instead of controlling maximum leaf number, we tried maximum tree depth of 4, 6, 8, 10 and use 4 times number of trees in other 3 packages. Since CatBoost in \textit{SymmetricTree} mode uses simpler tree structure, it convergences slower. Thus for CatBoost in \textit{SymmetricTree} mode, we use 2000 trees for \textit{learning\_rate} 0.1, 4000 trees for \textit{learning\_rate} 0.05 and 20000 trees for \textit{learning\_rate} 0.01. Subsampling of training data and features are \textbf{disabled} in all experiments. For the accuracy and convergence rate experiments, we use the \textit{Ordered} mode of CatBoost for better accuracy. And for the training time experiments, we use \textit{Plain} mode since it is much faster. Other parameters are left as default.  

For XGBoost, LightGBM and CatBoost, we directly pick up the best result on test set in the best iteration of the best hyperparameter setting. For GBDT-PL, we sample 20\% from training data for validation, and pick the iteration and hyperparameter setting performs best in the validation set, then report the corresponding accuracy on test data. 

The versions of python packages we used are LightGBM 2.1.0, XGBoost 0.81 and CatBoost 0.14.2.  
\begin{table}[ht!]
\centering
\captionof{table}{Parameter Settings for LightGBM}
\begin{tabular}{c|c}
\hline\hline
num\_leaves & 16, 64, 256, 1024\\
\hline
max\_bin & 63, 255, 1024\\
\hline
min\_sum\_hessian\_in\_leaf & 1.0, 100.0 \\
\hline
learning\_rate & 0.01, 0.05, 0.1\\
\hline
reg\_lambda & 0.01, 10.0 \\
\hline
min\_data\_in\_leaf & 0, 20 \\
\hline\hline
\end{tabular}
\label{param_lgb}
\end{table}
\begin{table}[ht!]
\centering
\captionof{table}{Parameter Settings for XGBoost}
\begin{tabular}{c|c}
\hline\hline
max\_leaves & 16, 64, 256, 1024\\
\hline
max\_bin & 63, 255, 1024\\
\hline
min\_child\_weight & 1.0, 100.0 \\
\hline
eta & 0.01, 0.05, 0.1\\
\hline
lambda & 0.01, 10.0 \\
\hline
grow\_policy & loss\_guided \\
\hline
tree\_method & hist\\
\hline\hline
\end{tabular}
\label{param_xgb}
\end{table}
\begin{table}[ht!]
\centering
\captionof{table}{Parameter Settings for CatBoost (\textit{SymmetricTree} mode)}	
\begin{tabular}{c|c}
\hline\hline
depth & 4, 6, 8, 10\\
\hline
border\_count & 63, 128, 255\\
\hline
min\_data\_in\_leaf & 1, 100 \\
\hline
learning\_rate & 0.01, 0.05, 0.1\\
\hline
l2\_leaf\_reg & 0.01, 10.0 \\
\hline
grow\_policy & SymmetricTree \\
\hline
leaf\_estimation\_method & Newton\\
\hline
random\_strength & 0.0\\
\hline
bootstrap\_type & No\\
\hline\hline
\end{tabular}
\label{param_cat_sym}
\end{table}
\begin{table}[ht!]
\centering
\captionof{table}{Parameter Settings for CatBoost (\textit{Lossguide} mode)}	  
\begin{tabular}{c|c}
\hline\hline
max\_leaves & 16, 64, 256, 1024\\
\hline
border\_count & 63, 128, 255\\
\hline
min\_data\_in\_leaf & 1, 100 \\
\hline
learning\_rate & 0.01, 0.05, 0.1\\
\hline
l2\_leaf\_reg & 0.01, 10.0 \\
\hline
grow\_policy & Lossguide \\ 
\hline
leaf\_estimation\_method & Newton\\
\hline
random\_strength & 0.0\\
\hline
bootstrap\_type & No\\
\hline\hline
\end{tabular}
\label{param_cat_sym}
\end{table}
\begin{table}[ht!]
\centering
\captionof{table}{Parameter Settings for GBDT-PL}	
\begin{tabular}{c|c}
\hline\hline
max\_leaf & 16, 64, 256, 1024\\
\hline
max\_bin & 63, 255\\
\hline
min\_sum\_hessian\_in\_leaf & 1.0, 100.0 \\
\hline
learning\_rate & 0.01, 0.05, 0.1\\
\hline
l2\_reg & 0.01, 10.0 \\
\hline
grow\_by & leaf \\
\hline
leaf\_type & half\_additive \\ 
\hline
max\_vars & 5 \\  	
\hline\hline
\end{tabular}
\label{param_cat_sym}
\end{table}

\end{document}